\documentclass[11pt,authoryear,review]{elsarticle}
\usepackage[final,commandnameprefix=ifneeded]{changes}

\usepackage[bookmarks=true, colorlinks=true]{hyperref}

\usepackage{url}
\usepackage{multirow}
\usepackage{multicol}
\usepackage{booktabs}
\usepackage{graphicx}
\usepackage{lscape}
\usepackage{amsfonts}
\usepackage{xcolor}
\usepackage{siunitx} 
\usepackage{float}
\usepackage{dsfont}
\usepackage{url}
\usepackage{ulem}
\usepackage{comment}
\usepackage{algorithm2e}
\usepackage{threeparttablex}
\usepackage{pdflscape}
\usepackage{rotating}
\usepackage{pdflscape}
\usepackage{amssymb,amsmath}
\usepackage{subfigure}

\usepackage{lineno}
\usepackage{tabularx}
\usepackage{longtable}
\usepackage{xltabular}
\usepackage{caption}
\usepackage{array}
\newcolumntype{C}[1]{>{\centering\arraybackslash}p{#1}}

\journal{Remote Sensing of Environment}

\begin{document}
	
\begin{frontmatter}

\title{Automated forest inventory: analysis of high-density airborne LiDAR point clouds with 3D deep learning}


\author[label1]{Binbin Xiang\corref{CorAuthor}}
\cortext[CorAuthor]{Corresponding author.}
\ead{binbin.xiang@geod.baug.ethz.ch}
\author[label2]{Maciej Wielgosz}
\author[label1]{Theodora Kontogianni}
\author[label1]{Torben Peters}
\author[label2]{Stefano Puliti}
\author[label2]{Rasmus Astrup}
\author[label1]{Konrad Schindler}
\address[label1]{Photogrammetry and Remote Sensing, ETH Z{\"u}rich, 8093 Z{\"u}rich, Switzerland\\}
\address[label2]{Norwegian Institute of Bioeconomy Research (NIBIO), 1433 {\AA}s, Norway\\}

\begin{abstract}
Detailed forest inventories are critical for sustainable and flexible management of forest resources, to conserve various ecosystem services. Modern airborne laser scanners deliver high-density point clouds with great potential for fine-scale forest inventory and analysis, but automatically partitioning those point clouds into meaningful entities like individual trees or tree components remains a challenge. The present study aims to fill this gap and introduces a deep learning framework\added[id=R1]{, termed ForAINet,} that is able to perform such a segmentation across diverse forest types and geographic regions. From the segmented data, we then derive relevant biophysical parameters of individual trees as well as stands. The system has been tested on FOR-Instance, a dataset of point clouds that have been acquired in five different countries using surveying drones. The segmentation back-end achieves over 85\% F-score for individual trees, respectively over 73\% mean IoU across five semantic categories: ground, low vegetation, stems, live branches and dead branches. Building on the segmentation results our pipeline then densely calculates biophysical features of each individual tree (height, crown diameter, crown volume, DBH, and location) and properties per stand (digital terrain model and stand density). Especially crown-related features are in most cases retrieved with high accuracy, whereas the estimates for DBH and location are less reliable, due to the airborne scanning setup.
\end{abstract}


\end{frontmatter}


\section{Introduction}
\label{Sec:Introduction}

Forests offer multiple ecosystem functions and services such as timber production, carbon sequestration, valuable habitats to safeguard biodiversity, and recreation~\citep{Li2023AprDeep,Maes2023Jun22Accounting}. Forest inventories are intended to provide the required information to support sustainable, multi-functional forest management, so as to ensure the continued provision of those services. Up-to-date and fine-scale (i.e., tree level) information represents a cornerstone for the transition towards a small-scale, targeted, adaptive, and multi-functional management. 

Collecting and maintaining forest inventories was one of the first applications to adopt airborne laser scanning (ALS) as an operational tool, at local~\citep{ErikNaesset2004Laser}, regional and national scales~\citep{Kangas2018Remote}. The two main approaches to extract forest inventory data from Light Detection and Ranging (LiDAR) point clouds are the \emph{area-based approach}~\citep[ABA,][]{Naesset2002Predicting} and \emph{individual tree detection}~\citep[ITD,][]{Hyyppa2001segmentation}. ABA operates by estimating statistical indicators from the points within a given area (e.g., a grid cell). In contrast, ITD (also referred to as \emph{single tree inventory}) relies on the detection and segmentation of individual trees in the point cloud~\citep{Persson2002Detecting}\added[id=R1]{, and consequently requires data at higher resolution~\citep{COOMES201777}.} Several methods have been developed for ITD, whose accuracies differ depending on the forest conditions and the point density~\citep{Vauhkonen2011Comparative,Kaartinen2012International}. Most ITD algorithms operate on a rasterised canopy height model (CHM) derived from the highest ALS returns, representing the canopy top layer. Consequently, they are limited to the dominant canopy trees and tend to miss an important share of intermediate and understory trees. Despite the large body of research on the topic, the operational uptake of ITD inventories has so far been limited, due to (1) the bias introduced by the relatively low detection rates, caused by the omission of non-dominant trees, and (2) a lack of transferability across different forest conditions and datasets.

\replaced[id=R1]{Traditionally, low-resolution ALS-based inventories have been based on rather sparse point clouds (0.5\,--\,5\,pts/m\textsuperscript{2}) for ABA, and slightly denser for ITD (10\,--\,20\,pts/m\textsuperscript{2}).}{Traditionally, ALS-based inventories have been based on point clouds with relatively low densities (0.5\,--\,5\,pts/m\textsuperscript{2}) for ABA, and slightly higher densities for ITD (10\,--\,20\,pts/m\textsuperscript{2}).} With the rapid development of sensor technology, very high density ALS (ALS-HD) has become available\replaced[id=R1]{, with point cloud densities ranging from 500 up to 10,000\,pts/m\textsuperscript{2}, offering }{and offers} new opportunities~\citep{Kellner2019New,puliti_et_al_2023}. \replaced[id=R1]{While also collected with airborne platforms such as drones (as in the present study) or helicopters~\citep{Hyyppae2022Direct,Persson2022Twophase}, ALS-HD data differ from conventional ALS data in that they are acquired from lower altitude (and at lower flight speed), and usually with larger overlap between adjacent flight strips. Despite covering smaller geographical regions (e.g., forest stands), ALS-HD represents an emerging alternative data source for ITD inventories due to their (for airborne data) unprecedented level of detail, closing the gap between large-area ALS and high-resolution terrestrial laser scanning (TLS).}{The deployment of unmanned aerial vehicles or helicopters\mbox{~\citep{Hyyppae2022Direct,Persson2022Twophase}} as carriers for laser scanning now enables the acquisition of point clouds with densities of 500 up to 10,000 pts/m\textsuperscript{2}. These provide an unprecedented (for airborne data) level of detail and close the gap between large-area ALS and high-resolution terrestrial laser scanning (TLS).}

The increased amount of 3D structural detail, with distinctively visible individual trees throughout the vertical canopy profile, has sparked renewed interest in the pursuit of single tree inventories, and possibly even completely airborne inventories based on dense retrievals of the variables of interest from the point clouds~\citep{Liang2019Forest,Puliti2020Estimation}. In addition to the measurement of common forestry variables like the diameter at breast height~\citep[DBH,][]{Kuzelka2020Very} and tree biomass~\citep{Brede2019Nondestructive}, recent research has highlighted the potential to derive a suite of useful tree analytics and insights from per-tree ALS-HD data, such as species~\citep{Hakula2023Individual}, stem curve~\citep{Hyyppae2022Direct}, growth trajectory~\citep{puliti_et_al_2023}, and breeding-related traits~\citep{Toit2023Modelling}. The promise of high-resolution scans for ITD is evident, but there is a bottleneck: to measure the properties of each tree, one first needs a sufficiently accurate and transferable method to segment the raw scan data into individual trees.

Parallel to the rise of high-density ALS, we have seen an explosion of interest in deep machine learning algorithms for processing 3D point clouds in general, and forest scans in particular (Table~\ref{tab:RepresentativeMethods}). Over the last decade, the superior capabilities of deep neural networks when it comes to representing and analysing structured data have brought about transformative changes across a range of science and engineering disciplines~\citep{Wang2023Aug2Scientific}. Deep learning also holds great promise for tree segmentation, and indeed there has been a recent trend to adopt it for the ongoing efforts towards automated forest inventories~\citep{Xi2018Aug2Filtering,Chen2021Jan21Individual,Krisanski2021Apr7Sensor,Sun2022Jun14Individual,Chang2022Oct6Two,Wang2023Feb13Tree,Zhang2023Jun5Towards,Kim2023Jun5Automated,Jiang2023Jun25LWSNet, Wielgosz2023Jul27P2T,Straker2023Instance}.\added[id=R1]{ The review in Table~\ref{tab:RepresentativeMethods} indicates that most methods either apply 2D CNNs on CHM or use PointNet++~\citep{Qi2017PointNetDH} to directly process point clouds. Networks that operate on the CHM share the limitations of traditional 2.5D ITD algorithms, i.e., they disregard understory trees. The seminal PointNet++ architecture does consider all scan points, including those below the main canopy. But new network designs, particularly sparse 3D convolutional networks~\citep{Choy2019June4D}, offer higher accuracy as well as computational advantages, making them more suitable for high-density point clouds.}

{
\begin{landscape}
\begin{table}[htbp]
\centering
\tiny
\caption{Representative deep learning based methods for single and multiple tasks to segment trees based on point clouds}
\label{tab:RepresentativeMethods}
\begin{tabular}{p{2.5cm}p{7cm}p{8.5cm}p{2.5cm}}
\hline
{\bf Task} &
  {\bf Highlights of method} &
  {\bf Remarks} &
  {\bf References} \\ \midrule
\multicolumn{4}{l}{\itshape Single task: individual tree crown segmentation (main output is individual tree crown delineation or crown width)} \\
\multirow{6}{*}{CHM based} &
  (1) Deep learning method to recognize trees and then use height-related gradient information to accomplish individual tree crown delineation. &
  Data collected by Unmanned aerial vehicle (UAV) LiDAR in Chizhou City, China; Data not open; Code not open. &
  \citep{Chen2021Jan21Individual} \\
 &
  (2) Deep learning object detection algorithm to identify individual tree crowns in height maps generated from point cloud. &
  Data collected by ALS in Nanjing, China; Data open (Contact the author); Code not open. &
  \citep{Sun2022Jun14Individual} \\
 &
  (3) CHM based segmentation by using YOLOv5 network.  & FOR-instance dataset; Code open.
   &
  \citep{Straker2023Instance} \\ \midrule
\multicolumn{4}{l}{\itshape Single task: individual tree segmentation (main output are point-wise instance IDs)} \\
CHM based &
  (4) RandLAnet to remove non tree points. 2D detection based deep learning for tree instance segmentation and post processing refinement. &
  Data collected by TLS in Evo, Finland and Guigang, China; Other output is tree position, tree height, tree DBH, crown diameter; Data not open; Code not open. &
  \citep{Chang2022Oct6Two} \\
Point based &
  (5) Top-down instance segmentation deep learning and self-adaptive mean shift clustering. &
  Data collected by ALS in Washington, U.S.\ and Bretten, Germany; Data open; Code not open. &
  \citep{Zhang2023Jun5Towards} \\ \midrule
\multicolumn{4}{l}{\itshape Single task: individual tree component segmentation (main output are point-wise semantic labels)} \\
\multirow{9}{*}{Deep learning based} &
  (6) FCN for classify wood, branch and others. &
  Data collected by TLS in Canada; Data not open; Code not open. &
  \citep{Xi2018Aug2Filtering} \\
 &
  (7) Classify terrain, vegetation, coarse woody debris (CWD) and stem based on PointNet++. &
  Data collected by TLS, ALS, MLS and UAV-based aerial photogrammetry (UAS\_AP) in Australia and New Zealand; Other output is DTM; Data open (contact the author); Code open. &
  \citep{Krisanski2021Apr7Sensor} \\
 &
  (8) Classify ground, understorey, tree stem, tree foliage based on RandLA-Net. &
  Data collected by backpack laser scanner; Data not open; Code not open. &
  \citep{Kaijaluoto2022Semantic} \\
 &
  (9) Similar to (7). PointNet++ model for segmenting the canopy, trunk, and branches of tree. &
  Data collected by TLS in Korea; Data not open; Code not open. &
  \citep{Kim2023Jun5Automated} \\
 &
  (10) Similar to (7). Leaf-wood separation network. &
  Data collected by TLS in eastern Cameroon; Data open; Code not open. &
  \citep{Jiang2023Jun25LWSNet} \\
Hybrid features based &
  (11) Recurrent Neural Networks (RNNs) that directly estimates the geometric parameters of individual tree stems. &
  Data collected by ALS and TLS in Australia and New Zealand; Data not open; Code not open. &
  \citep{Wang2023Feb13Tree} \\ \midrule
\multicolumn{4}{l}{\itshape Multiple tasks} \\
 &
  (12) Step by step methods. At first ground remove; raster based faster RCNN for individual tree segmentation; 3D FCN for stem segmentation. &
  Data collected by TLS in Australia; Tasks include DTM generation, individual tree segmentation, stem points extraction and stem reconstruction; Data not open; Code not open. &
  \citep{Windrim2020May6Detection} \\
 &
  (13) PointNet++ for semantic segmentation. The remaining vegetation points are assigned to be the same tree as the nearest cylinder measurement in X and Y coordinates. &
  Data collected by TLS in Western Australia; Tasks include DTM generation, semantic segmentation (terrain, vegetation, CWD, stems), individual tree segmentation and tree attributes calculation (height, DBH); Data not open; Code open. &
  \citep{Krisanski2021Nov16Forest} \\
 &
  (14) PointNet++ semantic segmentation; Graph-based approach for instance segmentation. &
  Data collected by MLS in south-east Norway; Tasks include semantic segmentation (vegetation, terrain, stem, CWD), individual tree segmentation and derivation of attributes (height); Data not open; Code open. &
  \citep{Wielgosz2023Jul27P2T} \\ \bottomrule
\end{tabular}
\end{table}
\end{landscape}
}

Past research has predominantly focused on semantic segmentation tasks such as segmenting leaf and wood points or separating points on trees from those on understory~\citep{Krisanski2021Apr7Sensor}. Comparatively few studies look at tree instance segmentation~\citep{Windrim2020May6Detection,Krisanski2021Nov16Forest}, and those mostly rely on sequences of ad-hoc steps (possibly including deep learning modules), which makes it difficult to transfer them to a new location or sensor setup, due to the effort needed to correct set the hyper-parameters for every step of the sequence, including frequent dependencies between the settings of different steps~\citep{Wielgosz2023Jul27P2T}. 

Most previous studies about the analysis forest scans have focused on one specific sub-task, for instance only the localisation of individual trees, or the segmentation of tree crowns, or the segmentation of a given, individual tree into its components (Table~\ref{tab:RepresentativeMethods}). Individual tree localisation aims to identify a representative spatial location, typically the centre coordinates of the stem or crown~\citep{Zhang2019Jan16ANovel,PaulaPires2022Mar18Individual}. Generally, the analysis of (static or mobile) TLS data tends to extract stem centres, and stem attributes like the  DBH~\citep{Zhang2019Jan16ANovel,PaulaPires2022Mar18Individual}; whereas in ALS data (from drones as well as manned aircraft) it is common to look for the crown centre~\citep{Zoerner2018Nov13LiDAR}. Crown segmentation aims to delineate individual tree crowns, often based on 2D projections onto the ground, and to derive crown attributes such as the area, width or overlap~\citep{Strimbu2015Mar18graph,Michele2016Apr5Tree}. Individual tree segmentation refers to finding all points that belong to the same tree instance (including crown, stem and isolated branches)~\citep{Dai2018Aug17new,Yan2020Feb5Self}, whereas tree component segmentation consists in predicting fine-grained semantic labels (e.g., wood, foliage, stem) for all points of an individual tree~\citep{Xu2021Jun18Separation,Xi2018Aug2Filtering}.

Clearly, many or all of the mentioned tasks must be solved to meet the requirements of an inventory that supports forest management. A few studies have proposed pipelines that greedily carry out several tasks step by step~\citep{Wielgosz2023Jul27P2T,Windrim2020May6Detection}, or have collected existing solutions for individual tasks into a common software package~\citep{Roussel2020Sep24lidR}. See Table~\ref{tab:RepresentativeMethods}. We note that different segmentation tasks (tree vs.\ non-tree, individual tree instances, per-instance components) are inherently related to each other. Solving them simultaneously may offer significant synergies and result in better performance, while at the same time reducing the computational overhead. On the other hand, such a shared model is particularly demanding with respect to the underlying representation, as it must be universal enough to support multiple different tasks.

\added[id=R1]{Data augmentation synthetically enhances the diversity of the training data and prevents overfitting, thus improving model accuracy without the need to collect additional field data~\citep{Chen2021Jan21Individual,Zhang2023Jun5Towards}. 
The methods in Table~\ref{tab:RepresentativeMethods} usually employ basic data augmentation techniques like rotation, mirroring and coordinate jittering~\citep{Chen2021Jan21Individual,Krisanski2021Apr7Sensor,Kaijaluoto2022Semantic}. 
Yet, we are not aware of data augmentation strategies that take into account the specific characteristics of LiDAR point clouds captured in forests.

Another significant barrier to the advancement of automated forest inventory tools has been the lack of large reference datasets. Table~\ref{tab:RepresentativeMethods} highlights the limited availability of open data that, moreover, often provide only semantic category labels but not tree instance labels. The scarcity of labeled data for supervised learning presents a bottleneck for the use of deep learning. The emergence of comprehensive labeled datasets like FOR-Instance is an important prerequisite to enable 3D deep learning in support of forest inventories.}

In the present work, we consolidate data-driven segmentation and structuring of forest point clouds into a single, integrated deep learning framework. We argue that a key ingredient for an efficient, automatic airborne forest inventory and management system is a comprehensive segmentation engine\added[id=R1]{, tuned specifically to the data characteristics of forest scans.} In this way, reference data for different tasks can be jointly used to supervise the model training, so as to obtain a generic, versatile representation of point patterns that are informative about trees. Besides its proven abilities in terms of representation learning, the deep learning paradigm also offers end-to-end training from raw point cloud data, and thus a single, small set of hyper-parameters to tune the model to new forest and/or sensor characteristics. In our view, the segmentation problem is the critical part of the analysis. Once it has been solved with good accuracy, biophysical variables like tree height, DBH and crown attributes can be extracted with dedicated geometric algorithms, thus ensuring transparency and interpretability of the retrieval system. To validate the proposed approach, in the present study we:

\begin{itemize}
\item Develop a deep network\added[id=R1]{, ForAINet,} that ingests ALS-HD data and jointly performs semantic segmentation at plot/stand level (dominant canopy trees / intermediate and understory trees / low vegetation / ground), tree instance segmentation, and semantic component segmentation at tree level (stem / live branches / dead branches).
\item Complement the segmentation back-end with geometric algorithms to retrieve several important biophysical attributes from the segmentation results, including both individual tree attributes (height, crown diameter, crown volume, DBH, location), and plot-level attributes (stand density and terrain model).
\item Experimentally validate the performance on a database containing multiple plots from forests in different geographic regions, with varying characteristics. We also compare to a widely used point cloud processing software, and find that the proposed deep learning approach greatly outperforms that baseline.
\end{itemize}

All data used in our study is publicly available, and our source code will be made available under a permissive license. We hope that our work, together with the growing access to high-density LiDAR observations, will contribute to accelerating the development of automatic forestry inventories.

\section{Materials and methods}
\label{Sec:Materialsandmethods}

\subsection{Dataset}
\label{Sec:DatasetDescription}

\begin{figure}
\centering
\includegraphics[width=\textwidth,keepaspectratio]{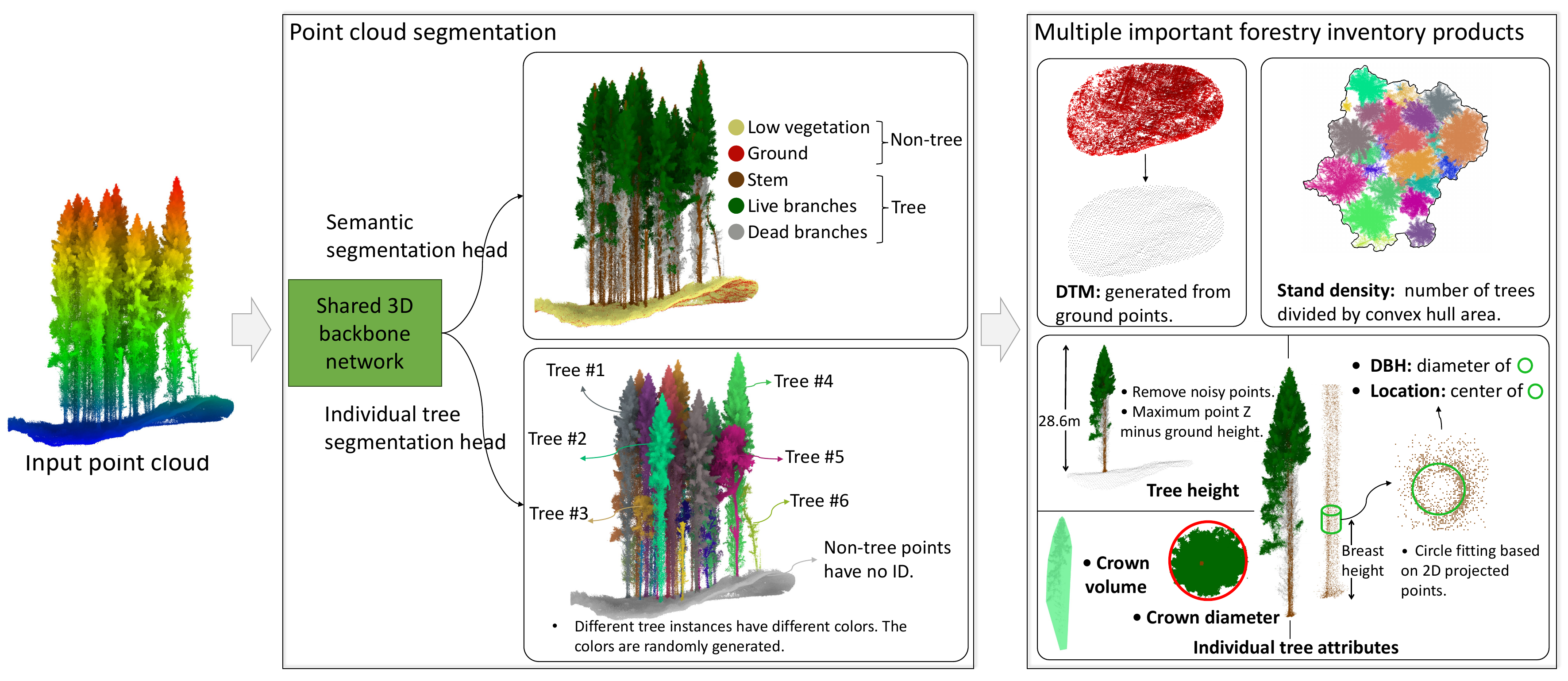}
\caption{Illustration of our segmentation and retrieval framework.\added[id=R1]{ It operates in two steps: The first step segments points into semantic categories as well as individual trees, for details see Figure~\ref{fig:pipeline} and Section~\ref{Sec:DeepLearningBasedFrameworkForMultipleSegmentationTasks}. The second step retrieves tree parameters and stand structure from the segmentation results, see Section~\ref{Sec:AutomatedQuantificationofImportantTreesFeaturesandStandStructure}.}}
\label{fig:workflow}
\end{figure}


\begin{table}[h]
\centering
\caption{\added[id=R1]{Characteristics of the FOR-Instance dataset in different geographic regions.}}
\label{tab:ForInstanceDataInfos}
\resizebox{\textwidth}{!}{
\begin{tabular}{p{1.4cm}p{4cm}p{4cm}p{4cm}p{0.6cm}p{0.6cm}p{0.6cm}p{1.2cm}p{3cm}p{1.5cm}}
\toprule
\multirow{2}{*}{\begin{tabular}[c]{@{}l@{}}Region\\ name\end{tabular}} &
  \multirow{2}{*}{Forest type} &
  \multirow{2}{*}{Name of tree species} &
  \multirow{2}{*}{Sensor} &
  \multicolumn{3}{c}{Number of plot} &
  \multirow{2}{*}{\begin{tabular}[c]{@{}l@{}}Number\\ of trees\end{tabular}} &
  \multirow{2}{*}{\begin{tabular}[c]{@{}l@{}}Average point\\ density (pts/m$^2$)\end{tabular}} &
  \multirow{2}{*}{Country} \\ \cmidrule(lr){5-7}
       &                                             &                      &                     & Train & Eval & Test &     &      &                \\ \midrule
CULS   & Coniferous dominated temperate forest       & Pinus silvestris     & Riegl VUX-1 UAV     & 1     & 1    & 1    & 47  & 2585 & Czech Republic \\
NIBIO &
  Coniferous dominated boreal forest &
  \begin{tabular}[l]{@{}l@{}}Picea abies (dominated)\\ Pinus Sylvestris (few)\\ Betula pendula (few)\end{tabular} &
  Riegl MiniVUX-1 UAV &
  8 &
  6 &
  6 &
  575 &
  9529 &
  Norway \\
RMIT   & Native dry sclerophyll eucalypt forest      & Eucalyptus pulchella & Riegl MiniVUX-1 UAV & 1     & 0    & 1    & 223 & 498  & Australia      \\
SCION  & Non-native pure coniferous temperate forest & Pinus radiata        & Riegl MiniVUX-1 UAV & 2     & 1    & 2    & 135 & 4576 & New Zealand    \\
TUWIEN & Deciduous dominated alluvial forest         & Deciduous species    & Riegl VUX-1 UAV     & 1     & 0    & 1    & 150 & 1717 & Austria        \\ \bottomrule
\end{tabular}%
}
\end{table}

The FOR-Instance dataset \citep{puliti2023forinstance} is a recently published, machine learning-ready collection of 3D point clouds with manually annotated per-point semantic class labels and tree IDs. The dataset also has DBH values acquired by field measurement for selected trees. The point clouds were collected with drones and helicopters equipped with survey-grade laser scanners (Riegl VUX-1 UAV and Mini-VUX) at 67 plots across various geographic regions and forest types. \replaced[id=R1]{For our study, points with reference label \emph{outpoints}, corresponding to incomplete or partially observed trees along plot borders, were removed due to the absence of instance labels for them.}{For our study, we removed points with reference label \emph{outpoints}, i.e., incomplete, partially observed trees along plot borders.} This leaves us with five semantic categories: \emph{low vegetation, ground, stem points, live branches} and \emph{dead branches}, see Figure~\ref{fig:workflow}. Points on stems, live branches and dead branches are considered tree points, the other two categories are non-tree points. The data is split into 42 training plots, 14 validation plots and 11 test plots, in such a way that all five geographic regions are present in both the training and the test portions.\footnote{The NIBIO2 region of FOR-Instance had to be excluded from the quantitative experiments, because it lacks DBH reference data, and because a significant number of understory trees are \replaced[id=R1]{not}{mot} annotated in the reference.}\added[id=R1]{ Table~\ref{tab:ForInstanceDataInfos} lists important characteristics of the FOR-Instance dataset, separately per geographic region. For further information about the data we refer the reader to \citet{puliti2023forinstance}.}

One specific challenge of the FOR-Instance data is a strong geographic imbalance: while the NIBIO forest region comprises 37 training plots, CULS, RMIT, and TUWIEN each only feature one training plot. The different regions vary greatly in terms of tree species composition and terrain. For example, the CULS region is predominantly made up of \emph{Pinus sylvestris}, whereas TUWIEN mainly consists of deciduous trees. These species differences result in a wide range of crown shapes and of the amount of live branches. Furthermore, the regions exhibit marked differences in point density. Particularly NIBIO is sparser than the other regions near and on the ground. Also, some regions (including NIBIO) have pronounced topography, whereas others are fairly flat. Finally, FOR-Instance features realistic, complex tree structures, including bent stems, occluded or sparsely sampled stems, and fallen trees lying on the ground.

\subsection{Deep learning framework for multiple segmentation tasks}
\label{Sec:DeepLearningBasedFrameworkForMultipleSegmentationTasks}

\begin{figure}
\centering
\includegraphics[width=\textwidth,keepaspectratio]{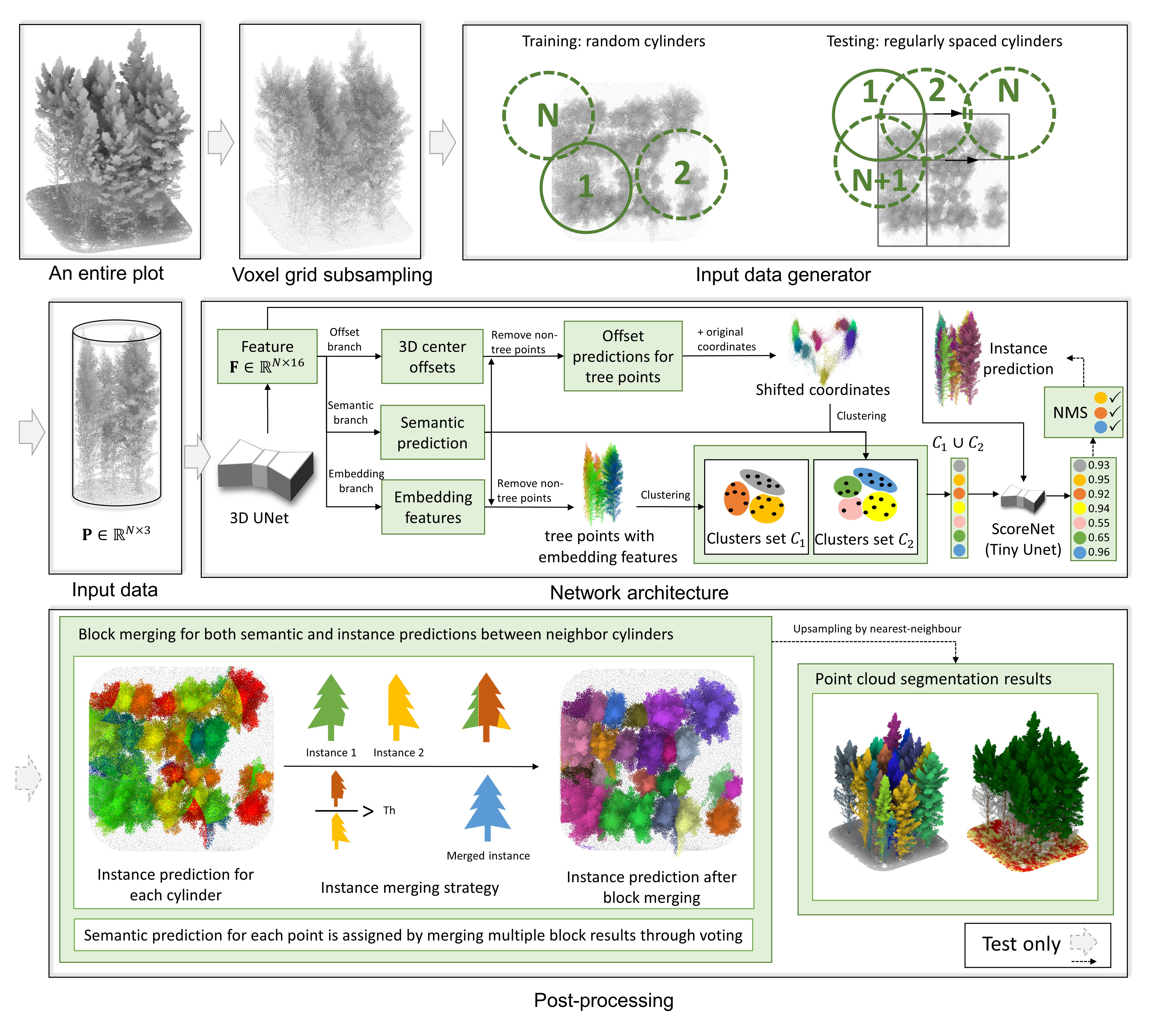}
\caption{\added[id=R1]{Illustration of the point cloud segmentation pipeline.}}
\label{fig:pipeline}
\end{figure}

Our forest inventory system consists of two parts, see Figure~\ref{fig:workflow}. First, a deep neural network labels the individual 3D points, simultaneously assigning semantic labels and instance IDs, using a shared feature extraction backbone\added[id=R1]{ (Figure~\ref{fig:pipeline})}.\added[id=R1]{ For brevity, we call that component ForAINet, short for \textbf{For}est \textbf{A}utomatic \textbf{I}nventory Neural \textbf{Net}work.} In a second step, learning-free geometric methods operate on the segmented data to retrieve a suite of forestry-related biophysical variables at the per-tree and per-stand levels. 

\subsubsection{\replaced[id=R1]{Data augmentation and balancing strategies}{Input data}}
\label{Sec:InputDataAugmentation}

The input to the segmentation network is a point set $P\in\mathbb{R}^{N \times C}$, where  $N$ is the number of points, and $C$ is the dimension of the per-point attributes. In our base setting, the attributes are only the $x$, $y$, $z$ coordinates, with the origin at the point cloud's centre of gravity. If further useful information is available, e.g., additional sensor readings or forest-specific hand-crafted descriptors~\citep{Wang2020May28Unsupervised}, it can be appended as additional attributes. For the presents study, we have explored the point-wise intensity, the return number and the scan angle rank, as given by the FOR-Instance dataset. Moreover, we have experimented with \replaced{traditional, hand-crafted}{manually extracted} geometric features derived from the eigenvalues of the local second moment matrix, including their sum, omnivariance, eigenentropy, anisotropy, planarity, linearity, surface variation, sphericity~\citep{Hackel2016FAST} and verticality~\citep{Guinard2017Weakly}\added[id=R1]{, c.f.~Section~\ref{Sec:CodeImplementation}}. For efficiency, the entire point cloud is voxel grid subsampled\added[id=R1]{ to a single, randomly chosen point per voxel, so as} to thin out overly dense regions and achieve a homogeneous (maximum) point density. The filter voxel size is set to 20$\times$20$\times$20$\,$cm\textsuperscript{3}, respectively 125\,pts/m\textsuperscript{3}.

Training samples are drawn following a class-balanced random sampling scheme: 3D points are sampled with probability $P_i$ inversely proportional to the square root of the class frequency $N_i$ ($P_i\propto\sqrt{1/N_i}$)\added[id=R1]{, where the subscript $i$ indicates the five different semantic classes}. The sampled point defines the axis of a vertical cylinder of fixed radius, and the set of all points within the cylinder forms one training sample. \added[id=R1]{In our implementation the radius is set to 8\,m, enough to ensure almost every tree in the FOR-Instance data is contained in a single cylinder (for other biomes this value may have to be adapted).}

In addition to the class-balanced sampling, we also experimented with three additional balancing strategies\added[id=R1]{, in order to determine and adopt the most effective strategy.} \emph{Class weighting} assigns weights inversely proportional to the square root of the per-class point count when computing the (semantic) segmentation loss, such that classes that are rare within a given sample receive higher weights. \emph{Height weighting} assigns higher weights to low points (small $z$-coordinate), to reflect the uneven scan density along the vertical. The height values are normalized by their mean and the weight is inversely proportional to the logarithm of that height. Finally, for more balanced sampling across different regions, we test a \emph{region weighting} scheme, where each geographic region is assigned a sampling probability, calculated from the region-wide class frequencies with the same inverse square root scheme as for class balancing.


For each sampled cylinder region, various data augmentation techniques are applied during training: additive Gaussian random noise on the point coordinates (jittering), random rotations around the cylinder axis, random anisotropic scaling by factors $s\in[0.9,1.1]$, and random reflection along the $y$-axis. In addition, \replaced[id=R1]{a subsampling strategy}{ dropout} is implemented for all input points: for every training sample, an unbiased coin flip (50\% probability) decides whether to apply \replaced[id=R1]{subsampling}{ dropout}; if yes, then 40\% of the points in the sample are randomly discarded. That procedure aims to boost robustness against occlusions and missing points, especially for parts obscured by foliage and for the part near the ground, which tends to be sparser in ALS point clouds. Moreover, \replaced[id=R1]{elastic deformation is applied, with the aim to enhance the recognition of curved tree stems. To that end, a field of Gaussian random deformations with magnitude $\in[0.4,1.6]\,$m is sampled on a regular grid of granularity $\in[0.2,0.8]\,$m. The field is then smoothed to ensure spatial coherence and applied to the point coordinates via trilinear interpolation, causing local bending of previously straight structures such as stems.
}

\begin{figure}
\centering
\includegraphics[width=\textwidth,keepaspectratio]{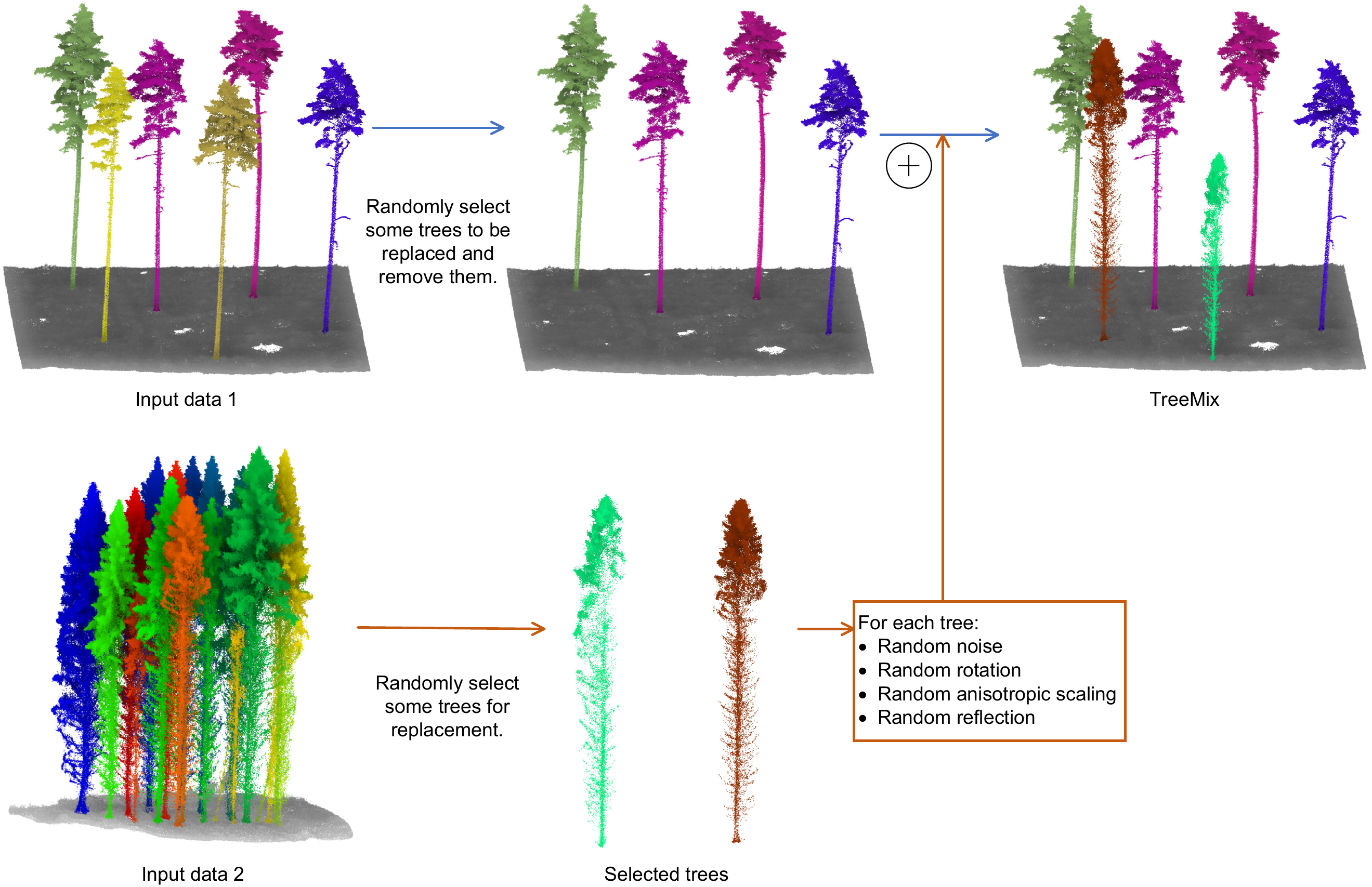}
\caption{Illustration of pipeline of TreeMix method for data augmentation.}
\label{fig:treemix}
\end{figure}

Inspired by the idea  behind Mix3D~\citep{ Nekrasov2021Mix3D} and CoSMix~\citep{ Saltori2022CoSMix}, we propose TreeMix, a dedicated data augmentation strategy for forest point clouds. As illustrated in Figure~\ref{fig:treemix}, TreeMix synthetically mixes two cylindrical training samples: tree instances in the target sample are randomly removed and replaced by with instances taken from the source sample, by copying them to the appropriate locations. To further enhance diversity, the newly added trees are augmented with random noise, rotation, reflection and scaling, as described above. A newly inserted tree is accepted only if it has low overlap with the points in the target sample (overlap below 10\%). This check ensures that synthetic samples comply with the logic of manually labeled forest point clouds, where different tree IDs do not overlap, since inter-penetrations between different canopies cannot be labeled precisely. By mixing trees from different locations within a plot, or even from different plots or countries, TreeMix increases the diversity of the training data.

\subsubsection{Network architecture}
\label{Sec:NetworkStructure}
The point cloud segmentation network consists of a shared 3D feature extraction backbone, followed by three parallel prediction heads\added[id=R1]{ (Figure~\ref{fig:pipeline})}. The first head is dedicated to semantic segmentation.\replaced[id=R1]{ To address the complexity of segmenting trees in densely packed forests, it proved beneficial to use two heads to}{The two others} extract two complementary embeddings of each point, which serve as the basis for tree instance clustering.

As feature extraction backbone we use a sparse 3D convolutional neural network (CNN), implemented in the MinkowskiEngine library~\citep{Choy2019June4D}. It offers a favourable trade-off between performance and computational cost~\citep{Xiang2023Review}. Essentially, the backbone is a 3D version of the U-Net architecture that processes voxelised point clouds with sub-manifold sparse convolutions, generating per-point feature vectors of length 16. These features are then passed to the three prediction heads.

The semantic segmentation branch estimates a semantic label for each point. First, the per-point features are fed through a multi-layer perception (MLP) to obtain 5-class semantic scores $F_s\in\mathbb{R}^{N \times 5}$. Those scores are further processed with two branches. One simply applies a \emph{softmax} activation to obtain 5-class probabilities for the point. The other has one more hidden layer, also followed by \emph{softmax} activation, to obtain binary tree/non-tree probabilities. Both outputs are trained with standard cross-entropy losses. Points labeled as non-tree are excluded from the subsequent individual tree segmentation.

The instance segmentation branch assigns point-wise individual tree IDs by clustering the outputs of the two corresponding prediction branches. One of them estimates, for each point, a 3D offset vector from the center of its tree instance, and is supervised with loss consisting of (i) the cosine distance between the true and predicted offset vectors\added[id=R1]{,  to align the directions of the predicted and true offset vectors, irrespective of object size and distance from the object center}; and (ii) the $L_1$ distance between their endpoints\added[id=R1]{ to directly minimize the deviation from the true offset}. The 3D centre offset\added[id=R1]{ has been} advocated by several previous studies~\citep{Jiang2020JunePointGroup,Chen2021Hierarchical,Vu2022SoftGroup,Zhong2022MaskGroup}\added[id=R1]{.}\deleted[id=R1]{ is essentially a learned version of the generalised Hough transform:} \replaced[id=R1]{A perfect prediction}{perfect predictions} would mean that the offsets contract each instance to a single point. The other branch maps each point to a 5-dimensional embedding space, such that points on the same tree instance form clusters in that space.\added[id=R1]{ Using 5 dimensions is in line with existing literature~\citep{Wang2019Associatively, Engelmann20CVPR,He2020LearningAM} and corroborated by our own experiments, where higher values did not produce significant accuracy gains.} It is supervised with a contrastive loss function that aims to minimise distances between points on the same tree, and to maximise distances between points on different trees~\citep{Wang2019Associatively}. Notably, the two additional dimensions of the embedding space make it possible to represent further object properties beyond 3D geometric offsets. \added[id=R1]{Empirically, using both the embedding and the direct offset prediction allows for to more accurate tree instance segmentation.}

The two embeddings are separately clustered to tree candidates \added[id=R1]{in unsupervised fashion}: offsets are explicitly added to the point coordinates and the resulting, shifted points are clustered with simple region growing\replaced[id=R1]{ as recommended by~\citep{zhao2021technical}}{\mbox{~\citep{Xiang2023Review}}}, with the threshold for the maximum distance\added[id=R1]{ set at 0.3 meters.} Clusters in the 5D embedding space, where distances do not have a direct geometric interpretation, are found with the mean-shift method\added[id=R1]{, as in several studies~\citep{Wang2019Associatively,Lahoud20193DIS}. Mean-shift~\citep{Comaniciu2002Mean} has only a single bandwidth parameter, which we always keep at an empirically determined value as 0.6, a setting we found to robustly work across city as well as forest scenes~\citep{Xiang2023Review,Xiang2023Towards}. In line with previous work, we find that conventional clustering as a post-process works better than learning cluster assignments within the neural network.}

The two (redundant) sets of tree instance candidates are merged and filtered with another small neural network, called ScoreNet~\citep{Jiang2020JunePointGroup}\deleted[id=R1]{)}, that predicts how well each instance candidate matches a ground truth (GT) tree\deleted[id=R1]{.} instance (i.e., ScoreNet~\citep{Jiang2020JunePointGroup}). That ScoreNet is a small 3D U-Net followed by max-pooling and a fully connected layer, and is trained to regress the maximal expected intersection over union (IoU) between an instance candidate and any of the actual trees. The candidates are then ordered by their score and pruned with greedy non-maximum suppression to obtain the final set of individual tree instances. All non-tree points are assigned the instance label $-1$. See~\citep{Xiang2023Towards} for further details.

At test time cylindrical blocks are not sampled at random, but on a regular $(x,y)$-grid, to ensure even coverage of the plot. Overlapping blocks are merged by re-assigning instance IDs such that they are globally unique, while greedily fusing instances that were split across more than one block~\citep{Xiang2023Towards}. We slightly modify the original block merging scheme and fuse the instances whose intersection has the highest overlap with the smaller instance (rather than with the union). After block merging the points discarded by the initial voxel grid subsampling are reinserted and labeled with the nearest-neighbour method, such that every observed scan point has a semantic class and an instance ID.

\subsection{Automated retrieval of tree parameters and stand structure}
\label{Sec:AutomatedQuantificationofImportantTreesFeaturesandStandStructure}

After segmenting a 3D point cloud, important individual tree attributes and stand structure characteristics can be extracted with straightforward geometric computations\added[id=R1]{ (see Figure~\ref{fig:workflow})}. These biophysical attributes are the actual variables of interest for the forest inventory, because they typically serve as input for management decisions and ecological analysis. Compared to field measurements, tree parameters retrieved from point clouds allow for complete and consistent coverage of much larger areas. For some parameters (e.g., tree height), LiDAR retrievals are also the most accurate method; while for others (e.g., DBH) they are not as accurate as field measurements, and one has to accept a lower reliability of the individual measurement in exchange for the better coverage.
In the present study, we extract the following exemplary forest attributes: at plot level, a digital terrain model (DTM) of the plot and its stand density; and for each individual tree the height, crown diameter, crown volume, crown volume of live branches, DBH and location. 

The DTM (a raster of heights above some reference, in m) is routinely retrieved from LiDAR observations, due to the high cost of field surveys. In our study, the relevant points for DTM fitting have already been filtered by segmenting a separate ground class. Computing the DTM reduces to setting up a regular $(x,y)$-raster and interpolating the heights of the raster points. We choose a grid spacing of 0.5$\,$m$\times$ 0.5$\,$m. Due to the high point density, nearest neighbour interpolation is sufficient.

Stand density (trees/ha) is another important plot-level feature, reflecting a forest's growth state and influencing factors such as light availability, resource competition, and overall forest health~\citep{Liang2018Jul24International}. Stand density is calculated by counting the number of individual trees and dividing it by the surface area they cover. That area is estimated by projecting all tree points down to the $(x,y)$-plane and computing the area of the 2D convex hull around them (Figure~\ref{fig:workflow}).

The tree height (m above ground) is defined as the elevation difference between the highest point of a tree and the  ground level at the bottom of the same tree instance (Figure~\ref{fig:workflow}). The ground elevation is found by interpolating the DTM value at the $(x,y)$-location of the tree. For a more reliable estimate of the height, we found it advantageous to filter out potential outliers above the canopy (e.g., scanning artefacts, isolated, protruding twigs). To that end we cluster all points labeled as members of the individual tree with HDBSCAN~\citep{McInnes2017hdbscan} based on their 3D coordinates, find the dominant cluster, and return its highest point as the tree top.

There are several different ways to define the crown diameter (m), i.e., it may be the smallest enclosing circle, or the average between the major and minor axes of an enclosing ellipse. Some authors find the two longest perpendicular distances from the center line to the convex hull, compute their average and double it to approximate the diameter~\citep{Jan2017May43D,Chen2021Jan21Individual}. Others use the height and width of an axis-aligned bounding box~\citep{Sun2022Jun14Individual} or the mean radius of a circular ring that encloses the crown boundary (called a ``donut'')~\cite{Zhang2015Jun16Individual}. Here we simply project the crown points (live branches and dead branches) to the $(x,y)$-plane and find their smallest enclosing circle with Welzl's algorithm~\citep{Welzl19912005JunSmallest}. The diameter of this enclosing circle is our estimated crown diameter (Figure~\ref{fig:workflow}).

\added[id=R1]{Compared to the 2D crown diameter, crown volume more accurately describes the effective size of the tree's canopy, essential to assess photosynthesis potential and vitality. It enables the derivation of important ecological variables (e.g., growth) predictive of forest health and functionality.} To calculate the crown volume (m$^3$), we find all crown points belonging to the same individual tree and filter them with HDBSCAN as described above for the height estimation to discard isolated protrusions. For the dominant cluster we then calculate the 3D convex hull, i.e., the smallest convex polyhedron containing all points\deleted[id=R1]{.}\added[id=R1]{, as a simple and robust approximation of the volume, valid across varying point densities.} The volume of that polyhedron is our estimate of the crown volume. We compute two variants, one that includes live and dead branches and one that includes only live branches, where the latter is a more accurate description of the volume of the photosynthetic component of the tree and thus useful for its downstream use in modelling tree biophysical attributes such as tree volume and tree growth.

Traditional, manual field measurements of DBH (cm) are done with steel calipers. Often it is measured along two perpendicular directions and the two values are averaged~\citep{Liang2018Jul24International}. The breast height is usually set to 1.3$\,$m above ground. For automatic measurements of DBH based on the segmented point cloud, we follow the standard approach: find all stem points within a given height range around the breast height, project them onto the $(x,y)$-plane, and fit a circle to them (Figure~\ref{fig:workflow}). Specifically, we use a default height interval of $\pm$0.5$\,$m around the breast height, but increase that interval if necessary to ensure that at least 10 points are found. Additionally, we filter the projected 2D stem points with HDBSCAN clustering to remove isolated points due to stem segmentation errors. For additional robustness, the circle fit to the remaining points is performed with the RANSAC method~\citep{Fischler1981Random}. The circle diameter corresponds to the DBH, and the circle center determines the location of the tree, according to its definition as the stem center at breast height~\citep{Liang2018Jul24International}.

\subsection{Evaluation metrics}
\label{Sec:EvaluationMetrics}
\added[id=R1]{
For better readability, all metrics, including confusion matrices, mIoU scores and F-score, are shown as percentages, rounded to one decimal place.}
\subsubsection{Metrics for point cloud segmentation}
\label{Sec:MetricsPointCloudSegmentation}

Standard metrics are computed to evaluate semantic segmentation: confusion matrices, overall accuracy, mean per-class accuracy, and mean intersection-over-union (mIoU). Additionally, we assess the geometric accuracy of the estimated DTM.

{
\begin{table}
\centering
\caption{\added[id=R1]{Metrics used to evaluate individual tree segmentation. In machine learning, including computer vision, several of these metrics are also in common use but have different names.}}
\label{tab:MetricsForIndividualTreeDetection}
\small
\setlength{\tabcolsep}{4pt} 
\renewcommand{\arraystretch}{0.8} 
\begin{tabular}{@{}lll@{}}
\toprule
forestry term  & equation         & machine learning term \\ \midrule
\begin{tabular}[c]{@{}l@{}}Completeness\\ Tree detection accuracy (DA)\\ Producers's accuracy (PAdetect)\end{tabular} & $C$ = TP/N                & Recall (r)    \\ \hline
Omission error   & $e_{om}$ = FN/N        & 1-Recall           \\ \hline
Commission error & $e_{com}$  = FP/(TP+FP) & 1-Precision        \\ \hline
F-score          & F = 2rp/(r+p)    & F1-score           \\
\bottomrule
\end{tabular}%
\end{table}
}

When comparing the literature, we found that forest researchers~\citep{Yin2016Jul12How,Liang2018Jul24International} and machine learning professionals, including those specialising in computer vision~\citep{Gu2022Feb15review,Wang2019Associatively}, often use different names for the same quality metric. To make this explicit and avoid misinterpretations, we list them in Table~\ref{tab:MetricsForIndividualTreeDetection}. It shows that both fields differ in naming conventions while conveying similar evaluation criteria. In the Table~\ref{tab:MetricsForIndividualTreeDetection}, $N$ is the total number of reference trees, TP is the number of true positives (correctly detected trees),  FN is number of reference trees that are not detected, FP is number of false positives (detections not corresponding to any reference tree),
and TP+FP thus equals the total number of trees predicted by the system.

A subtle source of potential differences is the procedure used to match predicted trees to reference trees. In forest research this is frequently done based on geometric distances, i.e., each reference tree is matched to the detection that closest to it in terms of (stem) location~\citep{Yan2020Feb5Self,Huo2022Jan3Towards,Hao2022JanHierarchical}. Sometimes additional exclusion criteria are employed to ensure the detection is sufficiently close, has similar height, etc. A related variant is to consider a detection correct if its location falls within the crown boundary of the reference tree~\citep{Chen2022Jun10Individual}. Some studies on individual tree detection match by IoU score between projected 2D crown polygons~\citep{Dietenberger2023Tree}. 
In our study we follow the standard procedure in computer vision, which reflects the 3-dimensional nature of the problem. \replaced[id=R1]{Our method matches predicted tree instances to reference trees based on the IoU score between their respective point sets. This approach ensures robustness against variations in tree and stem densities across different plots.}{Detections are matched with reference trees at the point level, and matches are made based on the IoU score between the two point sets.} Predictions that do not reach an IoU of at least 0.5 with any reference tree are counted as false positives.

To go beyond verifying the presence of tree instances and evaluate how precisely they delineated, we introduce the coverage metric, which quantifies the agreement with the ground truth instance boundaries. Denoting the set of ground truth trees $\{I_i^\text{gt},i\in\{1,...,N_\text{gt}\}\}$ and the set of predicted trees as $\{I_j^\text{pre},j\in\{1,...,N_\text{pre}\}\}$, we compare the two sets on a per-instance basis. For each ground truth tree we find the predicted instance with the highest IoU score, formally:
\begin{equation}
    \text{maxIoU}(I_i^\text{gt}) = \max_{j=1}^{N_\text{pre}} \big(\text{IoU}(I_i^\text{gt},I_j^\text{pre}) \big)\;.
\end{equation}%
The coverage is then defined as the average over all reference trees:
\begin{equation}\label{Eq:Cov}
	\text{Cov}=\frac{1}{N_\text{gt}}\sum_{i=1}^{N_\text{gt}}\text{maxIoU}(I_i^\text{gt})\;.
\end{equation}

\subsubsection{Metrics for individual tree features and stand structure}
\label{Sec:MetricsOfIndividualTreeF eaturesandStandStructure}

For the DTM we use the two most common metrics~\citep{Liang2018Jul24International}: The \emph{root mean square error} (RMSE) measures the average (squared) vertical deviation from the reference. The \emph{coverage} derives indicates what portion of the plot area is covered by the estimated DTM: our nearest neighbour interpolation is based on a Voronoi tesselation~\citep{Shewchuk1996Triangle,Jonathan2002Delaunay}, as implemented in the \emph{pynn} function. It may thus happen that no DTM value can be derived, if the grid location falls outside the Voronoi cells.

As in most existing studies, the accuracies of the extracted tree height, crown diameter, crown volume, live crown volume and DBH are evaluated by scatterplots of matched individual trees~\citep{Gu2022Feb15review}. From those, several numerical metrics can be extracted: the \emph{slope} of the regression line, which should be close to 1; the correlation coefficient ($R^2$), which should also be close to 1; and the p-value, which should be close to 0, indicating a highly significant predictive relation between the estimates and the reference values. Moreover, we compute the RMSE of the estimated dimensions. 
Finally, for crown-related variables, we complement the RMSE with a relative version, RMSE\%, which is normalised by the \replaced[id=R1]{average}{avergae} reference value of the variable. In this way, the errors are set in relation to the absolute magnitude of the estimates: e.g., in a regions with small trees of $\approx$20$\,$m$^3$ crown volume, an error of $\pm$10$\,$m$^3$ is a lot, whereas in an area of tall trees with much higher volume, it is very little. For the tree location, we calculate the RMSE in $x$ and $y$ direction, respectively, and create positional accuracy plots to visualize the deviations from the reference coordinates.

\subsubsection{Evaluation for understory and suppressed trees}
\label{Sec:MetricsForConsideringSmallerTrees}

In many studies, trees smaller than a certain threshold (i.e., understory and suppressed trees) are excluded from the reference data collection. The criterion for what is considered ``too small" may differ across studies, typically it is determined by a lower bound on the DBH (e.g., 10$\,$cm is a typical value in temperate regions).
Nevertheless, omitting these smaller trees constitutes a loss of information about the understory, and introduces a bias when assessing the accuracy of remote sensing techniques: suppressed trees that are detected in the point cloud are counted as false positives, since there are no matching reference annotations. In the FOR-Instance data understory and suppressed trees are not individually identified in the reference data, but are categorized as ``low vegetation". Detected trees in the understory would thus be counted against our method, although the discrepancy is due to limitations of the ground truth, not the model.

We found that, due to the vertical airborne view through the canopy, ALS-based DBH estimates are relatively inaccurate compared to other tree variables. We therefore follow~\citep{Huo2022Jan3Towards} and use a height threshold instead: trees taller than 1/3 of the tallest tree in a plot are regarded as dominant, trees below that threshold constitute the understory. Only the dominant trees enter into the computation of the quality metrics in Section~\ref{Sec:MetricsPointCloudSegmentation}, whereas the understory can only be assessed by visual inspection, because there is no appropriate reference.
\subsection{Implementation details}
\label{Sec:CodeImplementation}

All code for our study was written in Python\added[id=R1]{, and is publicly available at \url{https://github.com/bxiang233/ForAINet}.} All experiments were conducted on a machine with 8-core Intel CPU, 8$\,$GB of memory per core, and one Nvidia Titan RTX GPU with 24$\,$GB of on-board memory. The \replaced[id=R1]{code for extracting hand-crafted geometric features}{manual feature extraction code} is a slightly modified version of the code published by~\cite{Guinard2017Weakly}. The point cloud segmentation network was constructed with the help of the Torch-Point3D library~\citep{Chaton2020}. All hyper-parameters for HDBSCAN filtering were determined by grid search on the validation set.

\section{Results}
\label{Sec:Results}
\subsection{3D point cloud segmentation}
\label{Sec:3DPointCloudSegmentation}

A series of experiments was conducted to examine whether various loss designs, weighting strategies and data augmentations could enhance segmentation performance for forest scenes. The quantitative results for individual tree segmentation and semantic segmentation, across different settings, are summarized in Table~\ref{tab:InsSemResults}. The classification results for each semantic class, as well as the binary tree vs.\ non-tree classification accuracy, are shown in Table~\ref{tab:SemResultsPerClass}. The basic setting is the panoptic segmentation network we developed in earlier work~\citep{Xiang2023Towards}, which we expanded to distinguish five semantic classes instead of only two (tree and non-tree). The other tested settings (Tables~\ref{tab:InsSemResults} and~\ref{tab:SemResultsPerClass}) are obtained by applying a single modification to the base setting.

\begin{table}[]
\centering
\caption{\added[id=R1]{Instance segmentation and semantic segmentation results under different settings. In each column, bold font indicates the best results.}}
\label{tab:InsSemResults}
\resizebox{\textwidth}{!}{%
\begin{tabular}{@{}lcccccccc@{}}
\toprule
\multirow{2}{*}{Settings}        & \multicolumn{5}{c}{Instance segmentation (\%)}  & \multicolumn{3}{l}{Semantic segmentation (\%)} \\ \cmidrule(l){2-6} \cmidrule(l){7-9} 
                       & Completeness & Omission error & Commission error & F-score  &  Cov    & oAcc  & mAcc           & mIoU  \\ \midrule
Basic setting          & 79.3        & 20.8          & 21.2           & 79.0    &  77.0    & 92.6 & 81.2          & 73.0 \\
+ binary semantic loss & 83.0        & 17.0          & 21.6           & 80.6    & \textbf{79.8}      & 92.5 & 80.4          & 72.4 \\
+ class weights        & 78.9        & 21.1          & 22.5           & 78.2   &   77.1    & 90.9 & \textbf{83.9} & 70.7 \\
+ height weights       & 80.8        & 19.2          & 21.4           & 79.7    &  78.0    & 91.8 & 78.4          & 70.4 \\
+ region weights       & 80.5        & 19.5          & 20.0           & 80.2     & 77.4    & 92.3 & 79.0          & 71.4 \\
+ intensity            & 78.9        & 21.1          & 14.7           & 82.0     &  76.7   & 93.0 & 78.9          & 72.7 \\
+ return number        & 81.7        & 18.3          & 13.5           & 84.1     &  78.2   & 92.9 & 81.6      & 73.7 \\
+ scan angle rank      & 81.7        & 18.3          & 18.5           & 81.6    &  79.0    & 93.2 & 81.8          & 74.3 \\
+ hand-crafted features    & 81.4          & 18.6          & 18.1 & 81.7 & 78.9&\textbf{93.6}    & 82.6   & \textbf{75.7}   \\
+ elastic distortion and \replaced[id=R1]{subsampling}{ dropout} & \textbf{83.3} & \textbf{16.7} & 16.5 & 83.4& 79.5 & 92.6             & 80.5   & 72.4            \\
+ TreeMix              & 82.0        & 18.0          & \textbf{11.7}  & \textbf{85.1} & 78.1  & 93.0 & 80.8          & 73.5 \\ \bottomrule
\end{tabular}%
}
\end{table}

\begin{table}[]
\centering
\caption{Per-class IoU values under different settings. Bold values in each column indicate the best results.}
\label{tab:SemResultsPerClass}
\resizebox{\textwidth}{!}{%
\begin{tabular}{@{}lccccccccc@{}}
\toprule
\multirow{2}{*}{Settings} &
  \multicolumn{6}{c}{Multi class semantic segmentation (\%)} &
  \multicolumn{3}{c}{Tree vs.\ non-tree (\%)} \\ \cmidrule(l){2-7} \cmidrule(l){8-10} 
 &
  Low veg. &
  Ground &
  Stem &
  Live branches &
  \multicolumn{1}{l}{Dead branches} &
  \multicolumn{1}{l}{mIoU} &
  Non-tree &
  Tree &
  mIoU \\ \midrule
Basic setting                    & 86.3          & 73.7          & 54.8          & 93.5 & 56.5 & 73.0 & 93.6          & 97.1          & 95.3          \\
+ binary semantic loss           & 84.8          & 70.2          & 54.3          & 94.1 & 58.8 & 72.4 & 96.0          & 98.1          & 97.1          \\
+ class weights                  & 78.9          & 65.1          & \textbf{58.2} & 93.4 & 57.7 & 70.7 & 95.9          & 98.1          & 97.0          \\
+ height weights                 & 84.0          & 67.6          & 50.6          & 93.3 & 56.6 & 70.4 & 92.6          & 96.6          & 94.6          \\
+ region weights                 & 85.5          & 72.5          & 48.7          & 93.4 & 56.7 & 71.4 & 95.0          & 97.7          & 96.4          \\
+ intensity                      & \textbf{89.3} & \textbf{78.8} & 50.1          & 93.4 & 51.9 & 72.7 & 92.9          & 96.7          & 94.8          \\
+ return number                  & 86.5          & 74.5          & 54.6          & 94.1 & 58.8 & 73.7 & 93.8          & 97.1          & 95.4          \\
+ scan angle rank                & 88.2          & 77.1          & 53.4          & 94.0 & 58.6 & 74.3 & \textbf{96.7}          & \textbf{98.4} & \textbf{97.6} \\
+ hand-crafted features &
  89.1 &
  78.7 &
  55.6 &
  \textbf{94.3} &
  \textbf{60.7} &
  \textbf{75.7} &
  94.2 &
  97.3 &
  95.7 \\
+ elastic distortion and \replaced[id=R1]{subsampling}{ dropout} & 85.8          & 73.9          & 50.8          & 93.9 & 57.8 & 72.4 & 92.6          & 96.6          & 94.6          \\
+ TreeMix                        & 87.7          & 76.0          & 52.3          & 94.0 & 57.4 & 73.5 & 95.9 & 98.1          & 97.0          \\ \bottomrule
\end{tabular}%
}
\end{table}

The tested setting can be categorized into three groups. The first group modifies the loss function or the sampling of training examples, i.e., the available input data is not modified, but used in a different manner. One way to do this is to add explicit supervision for the binary tree vs.\ non-tree task.\added[id=R1]{ It is implemented by adding a fully connected layer after the five-class semantic segmentation branch, and supervised with the binary classification loss.} At first glance this may appear superfluous, since the binary reference is derived by simply aggregating the fine-grained labels (stems, live branches and dead branches into the tree class, ground and low vegetation into the non-tree class). Still the aggregation adds information that otherwise is unavailable to the network, namely the hierarchical structure of the label space. E.g., as soon as a point is part of a tree, it can no longer be confused with the low vegetation, hence the classifier can invest its capacity into difficult decisions within the tree class, such as separating live from dead branches. Indeed, we empirically find that adding the binary labels increases the IoU for tree vs. non-tree by 1.8 percent points (Table~\ref{tab:SemResultsPerClass}). A more correct set of tree points, in turn, better supports instance discrimination, so that the F-score for individual tree segmentation also improves by 1.6 percent points (see Table~\ref{tab:InsSemResults}). The price to pay is a slightly higher confusion between the non-tree classes ground and low vegetation.

Also in this group fall different sampling and reweighting strategies. Sampling based on class frequency significantly improves the segmentation of the under-represented stem class (+3.4\,pp), and also of the dead branches (+1.2\,pp). However the associated accuracy loss for the ground ($-$2.3\,pp) hurts overall performance. Reweighting based on point height is meant to compensate the sampling bias of the ALS recording geometry, vertically through the canopy. Reweighting\deleted[id=R1]{ by} by the regional class distribution is a coarser version of class balancing, based on regional forest composition. The latter two strategies, however, both result in a moderate decline in IoU, despite small gains in regions with little training data.

 The second group of modifications concern the input data to the model. We tested different additional input observations beyond the point coordinates: LiDAR intensity, return number, scan angle rank, as well as hand-crafted statistical features as described in Section~\ref{Sec:InputDataAugmentation}. Compared to the basic setting, additional input features can to some extent enhance individual tree segmentation, see Table~\ref{tab:InsSemResults}. Among them, the return number stands out with a 5.1\,pp gain in F-score. For semantic segmentation, all additional features except for the intensity bring some improvement. In particular, the hand-crafted descriptors reach the highest semantic segmentation mIoU of 75.7$\%$. This may indicate that FOR-Instance is still too small to optimally train the network, as the network would, in principle, be able to (approximately) learn the extraction of the descriptors from raw data.

 The third group are tree-specific data augmentation strategies to synthetically increase the size and diversity of the training set. We have experimented with two approaches: elastic distortion with \replaced[id=R1]{subsampling}{ dropout}, and TreeMix. The idea behind the use of elastic distortion is to better account for curved and non-vertical stems, which are too rare to be learned well, but may still occur in the test data. The reasoning behind \replaced[id=R1]{subsampling}{ dropout}, i.e., removing points from the training examples, is to simulate points lost due to occlusion, scattering, etc., and make the network more robust against missing data. These low-level augmentation methods increase the instance segmentation F-score by 4.4\,pp, but slightly reduce the semantic segmentation performance. TreeMix is a more informed augmentation strategy, where samples are synthesised by mixing semantically meaningful entities (i.e., trees) from different samples. It achieves the best single tree segmentation result of 85.1\% F-score (+6.1\,pp), while also improving semantic segmentation.

\begin{figure}
\centering
\includegraphics[width=0.6\textwidth,keepaspectratio]{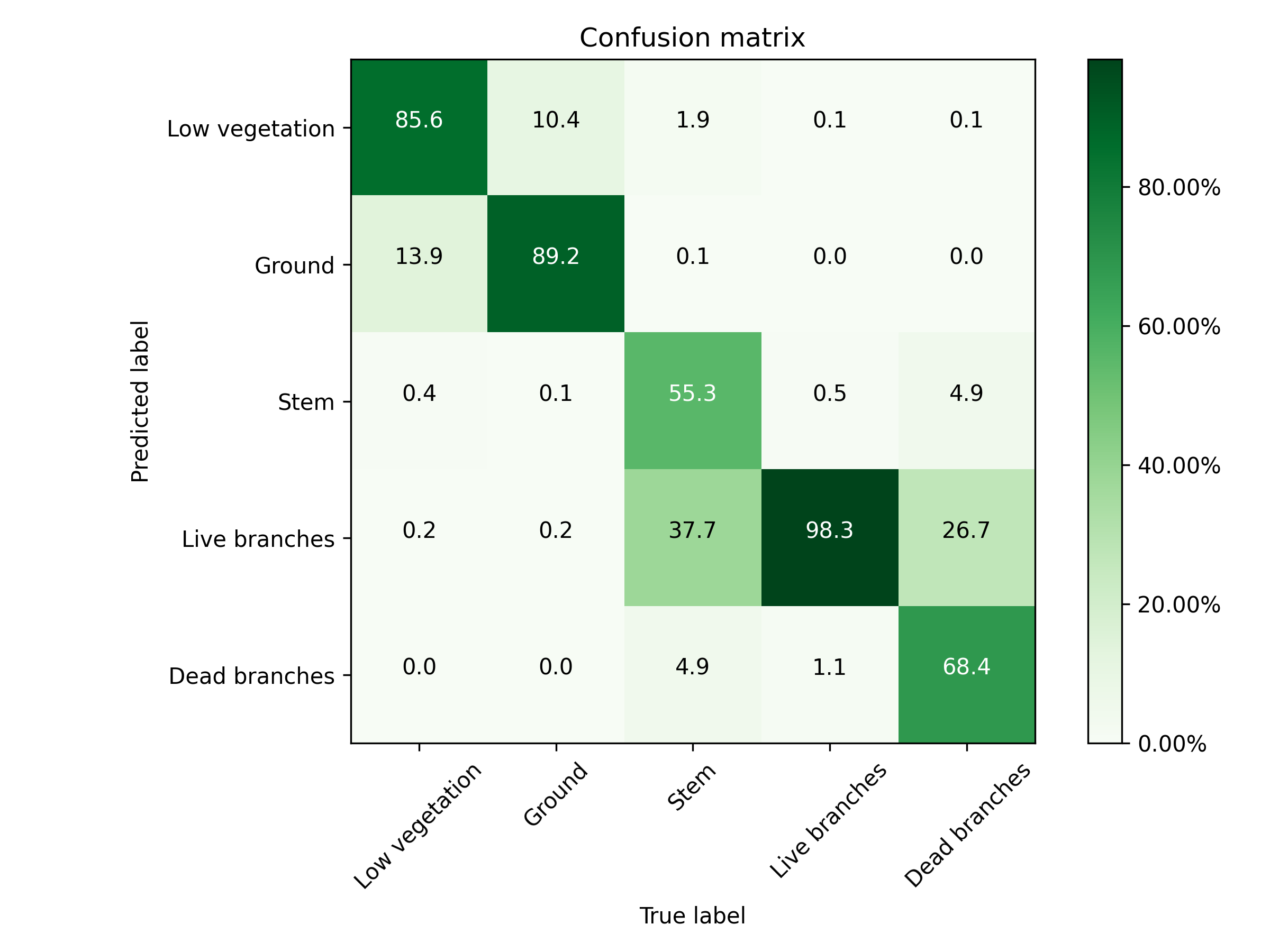}
\caption{\added[id=R1]{Confusion matrix between the five semantic categories, computed over all points of the test set.}}
\label{fig:confusionmatrix}
\end{figure}

Based on these results, we choose the network with TreeMix augmentation as the best configuration for the forest inventory task and conduct a series of detailed analyses with it. The confusion matrix (computed across all 11 plots of the test set) shows minimal confusion between tree (stem, live branches, dead branches) and non-tree points (low vegetation and ground), see Figure~\ref{fig:confusionmatrix}. As expected, confusions occur mainly between semantically similar categories, i.e., low vegetation and ground, respectively different tree components. Among them, the most significant confusion is that points on stems and dead branches are miss-classified as live branches.

\begin{figure}
\centering
\includegraphics[width=\textwidth,keepaspectratio]{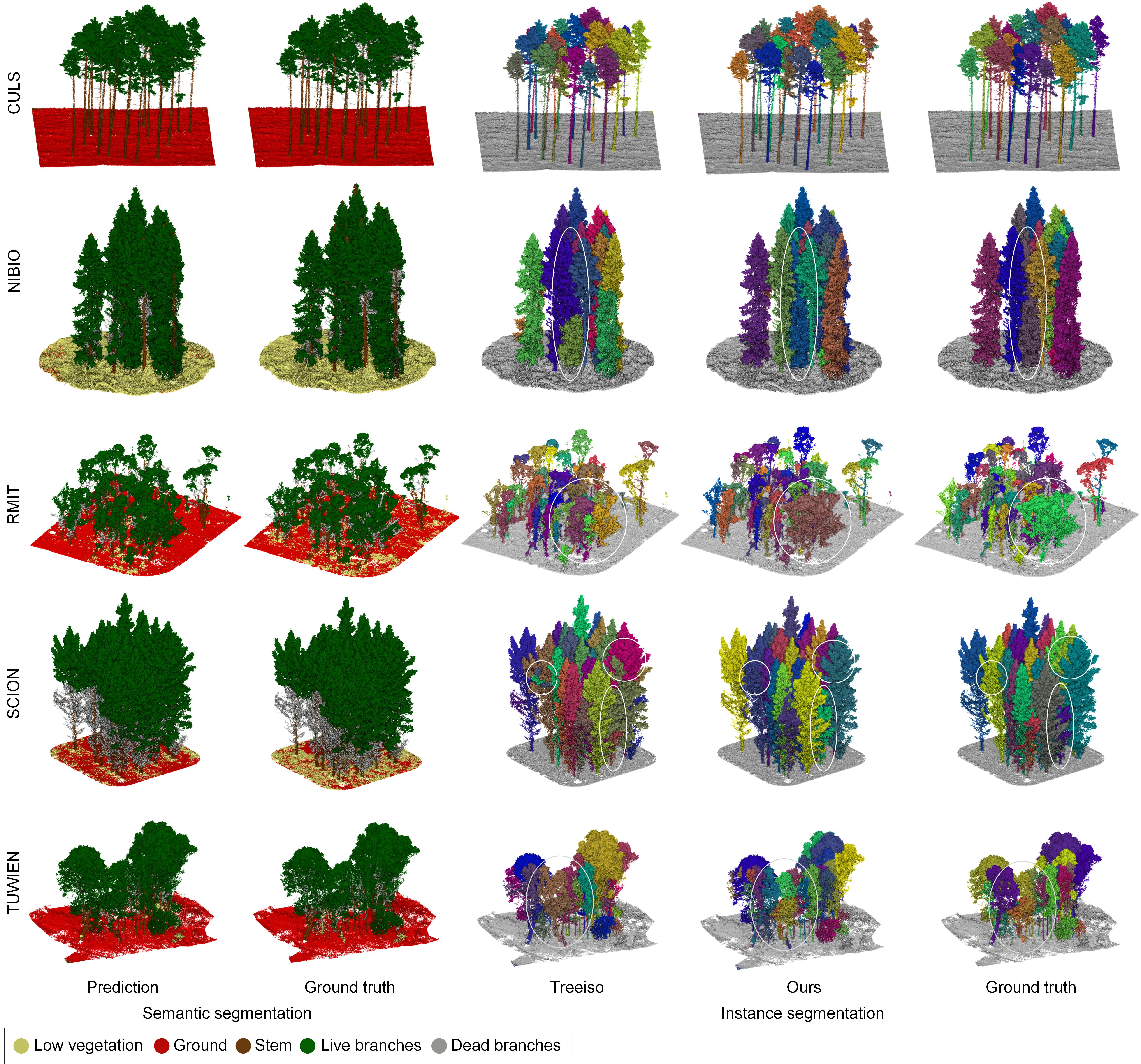}
\caption{Visual comparison between the proposed approach and Treeiso. The colours for the individual trees were assigned randomly. White ellipses mark noteworthy differences. Best viewed on screen.}
\label{fig:semanticandinstanceseg}
\end{figure}

Table~\ref{tab:IndividualTreeSegBest} presents individual tree segmentation results for each of the 11 test plots. Aside from Plot 8 (RMIT forest region) and Plot 11 (TUWIEN forest region), which show relatively poor results, all other plots achieve a F-scores above 82$\%$. A potential reason for the weaker performance on may be that RMIT has the lowest point density (on average, 498\,pts/m\textsuperscript{2}), and TUWIEN has the second-lowest one (1717\,pts/m\textsuperscript{2}); while all other forests have densities above 2585\,pts/m\textsuperscript{2}, with a maximum of 9529\,pts/m\textsuperscript{2} for NIBIO. The sparsity renders segmentation visibly more difficult, as can be seen in the third row of Figure~\ref{fig:semanticandinstanceseg}.

In addition, NIBIO, CULS, and SCION are pure coniferous or conifer-dominated forests. Coniferous trees are generally straighter and have slimmer crowns with smaller branches. On the contrary, RMIT is native dry sclerophyll Eucalypt forest, where eucalyptus trees have few, but thick branches and expansive crowns. TUWIEN is a deciduous-dominated alluvial forest, where trees have many branches and low crowns, varying greatly in size. Expectedly, our results show that segmenting tall and straight, slender trees is easier; even when they are located close to each other (Table~\ref{tab:IndividualTreeSegBest} and Figure~\ref{fig:semanticandinstanceseg}). Whereas for trees with dispersed crowns and complex branch structure, individual tree segmentation is more challenging. A further reason for this discrepancy the limited diversity of training data for the forest characteristics of RMIT and TUWIEN, which each only have a single plot of training data\added[id=R1]{ (Figure~\ref{tab:ForInstanceDataInfos})}.
 
\begin{table}[]
\centering
\caption{\added[id=R1]{Individual tree segmentation results for each plot in the test set.}}
\label{tab:IndividualTreeSegBest}
\resizebox{\textwidth}{!}{%
\begin{tabular}{@{}lcccrrrrrrcc@{}}
\toprule
\multicolumn{1}{l}{\begin{tabular}[c]{@{}c@{}}Plot ID\\ (Region)\end{tabular}} &
\multicolumn{1}{l}{\begin{tabular}[c]{@{}c@{}}reference\\ trees\end{tabular}} &
\multicolumn{1}{l}{\begin{tabular}[c]{@{}c@{}}trees with\\ DBH field data\end{tabular}} &
\multicolumn{1}{l}{\begin{tabular}[c]{@{}c@{}}detected\\ trees\end{tabular}} &  
\multicolumn{2}{c}{\begin{tabular}[c]{@{}c@{}}corr.\ detected trees\\ (Completeness)\end{tabular}} &
\multicolumn{2}{c}{\begin{tabular}[c]{@{}c@{}}omitted trees\\ (Omission \replaced[id=R1]{error}{ rate})\end{tabular}} &
\multicolumn{2}{c}{\begin{tabular}[c]{@{}c@{}}wrong detections\\ (Commission \replaced[id=R1]{error}{ rate})\end{tabular}} &
 \multicolumn{1}{l}{F-score} &
 \multicolumn{1}{l}{Cov}\\ \midrule
1 (CULS)  & 20 & 20 & 23 & \textcolor{white}{000000}20 & (100.0\%) & \textcolor{white}{00000}0 & (0.0\%)        & \textcolor{white}{000000}3 & (13.0\%)  & 93.0\% & 98.2\%\\
2 (NIBIO) & 37 & 26 & 26 & 26 & (70.3\%) & 11 & (29.7\%) & 0 & (0.0\%)        & 82.5\% & 62.4\%\\
3 (NIBIO) & 30 & 23 & 29 & 29 & (96.7\%) & 1 & (3.3\%)   & 0 & (0.0\%)        & 98.3\% & 88.3\%\\
4 (NIBIO) & 27 & 21 & 26 & 26 & (96.3\%) & 1 & (3.7\%)   & 0 & (0.0\%)        & 98.1\% & 84.6\%\\
5 (NIBIO) & 20 & 15 & 18 & 18 & (90.0\%)    & 2 & (10.0\%)     & 0 & (0.0\%)        & 94.7\% & 84.4\%\\
6 (NIBIO) & 28 & 18 & 29 & 26 & (92.9\%) & 2 & (7.1\%)   & 3 & (10.3\%)  & 91.2\% & 81.7\%\\
7 (NIBIO) & 19 & 19 & 18 & 16 & (84.2\%) & 3 & (15.8\%)  & 2 & (11.1\%)  & 86.5\% & 75.2\%\\
8 (RMIT) & 64 & 27 & 54 & 41 & (64.1\%) & 23 & (35.9\%) & 13 & (24.1\%) & 69.5\% & 60.6\%\\
9 (SCION) & 25 & 25 & 25 & 23 & (92.0\%) & 2 & (8.0\%)   & 2 & (8.0\%)   & 92.0\% & 86.7\%\\
10 (SCION) & 18 & 18 & 15 & 15 & (83.3\%) & 3 & (16.7\%)  & 0 & (0.0\%)        & 90.9\% & 79.4\%\\
11 (TUWIEN) & 35 & 35 & 37 & 25 & (71.4\%) & 10 & (28.6\%) & 12 & (32.4\%) & 69.4\% & 58.3\%\\ \bottomrule
\end{tabular}%
}
\end{table}

\subsection{Tree features and stand attributes}
\label{Sec:MeasurementofIndividualTreesFeaturesAndStandwiseAttributes}

\begin{figure}
\centering
\includegraphics[width=\textwidth,keepaspectratio]{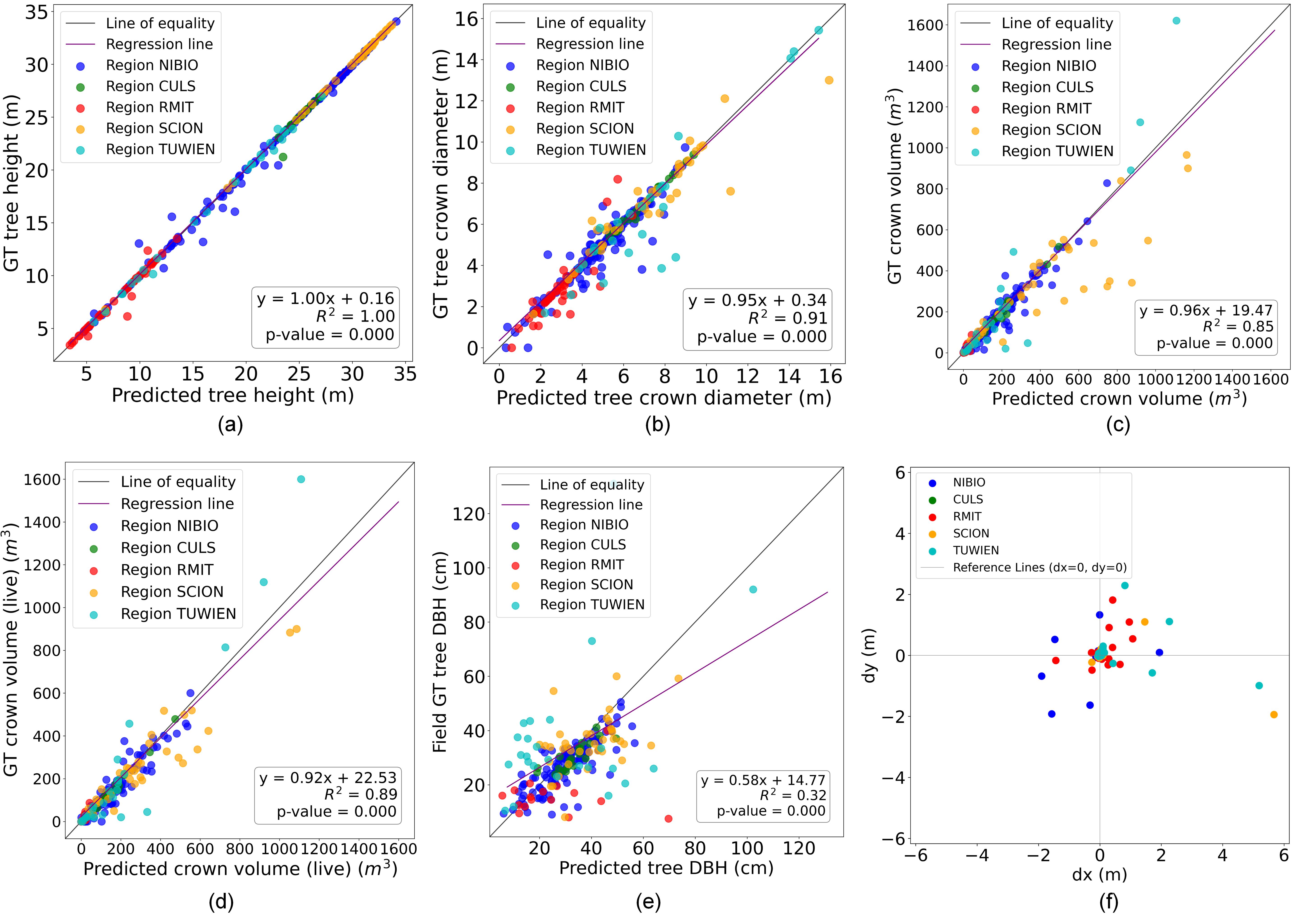}
\caption{(a)-(e) display scatterplots between the predicted and reference values for individual tree height, crown diameter, crown volume, crown volume of live branches, and individual tree DBH. (f) shows the positional accuracy (i.e., the deviations in $x$ and $y$ directions). Results for all matched trees from the five forest regions are shown in one graph.}
\label{fig:attributes_plot}
\end{figure}

\begin{figure}
\centering
\includegraphics[width=\textwidth,keepaspectratio]{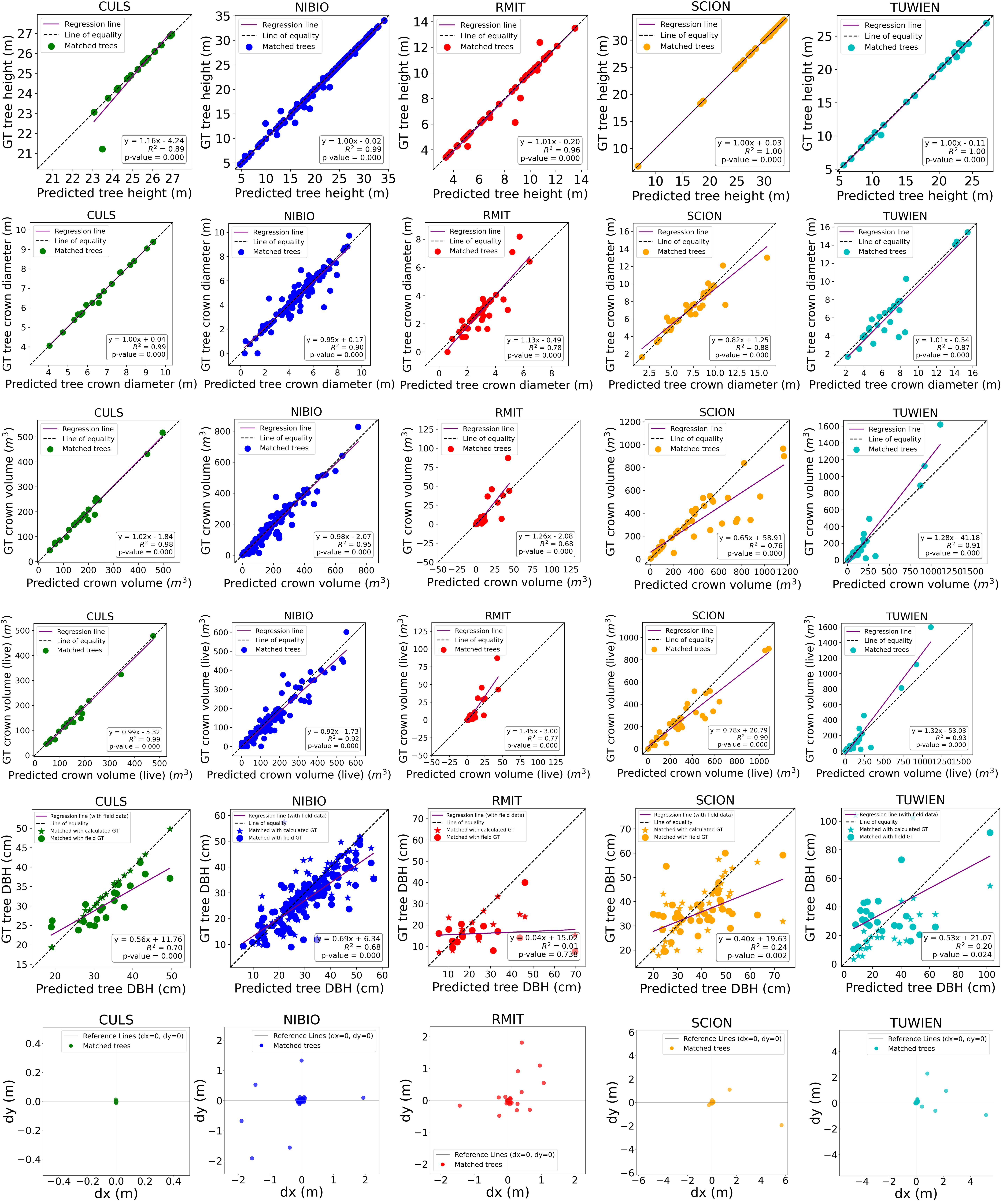}
\caption{Scatterplots of individual tree features (height, crown diameter, crown volume, crown volume of live branches, DBH) for matched trees, and positional accuracy plots.}
\label{fig:attributes_plot_per_region}
\end{figure}

Based on the segmentation results individual tree features and stand-wise attributes were calculated, as described in Section~\ref{Sec:AutomatedQuantificationofImportantTreesFeaturesandStandStructure}. Quality metrics obtained by comparing the predicted values to the reference are shown in Table~\ref{tab:attributesPerPlot}. Figure~\ref{fig:attributes_plot} shows the scatterplots for individual tree height, crown diameter, crown volume, crown volume of live branches, and individual tree DBH, and the positional accuracy plot for the matched trees from all five forest regions in one graph. Figure~\ref{fig:attributes_plot_per_region} shows scatterplots of matched individual trees and positional accuracy plots for each of the five forest regions.

As shown in Table~\ref{tab:attributesPerPlot}, the estimated DTM achieves high coverage ($>$98$\,$\%) and low RMSE value ($<$26$\,$cm) for all plots.\footnote{Note that the limiting factor for DTM is largely not measurement or fitting accuracy, but the considerable definition uncertainty of the forest floor.} Also tree height predictions are accurate, all RMSE$\%$ below 0.06 (see Table~\ref{tab:attributesPerPlot}). As shown in the scatterplots in Figure~\ref{fig:attributes_plot}(a) and in the first row of Figure~\ref{fig:attributes_plot_per_region}, estimated and reference tree heights correlate very well, across all forest regions. This finding is in line with a previous study, e.g., by~\cite{Wang2019Is}.

\begin{figure}
\centering
\includegraphics[width=\textwidth,keepaspectratio]{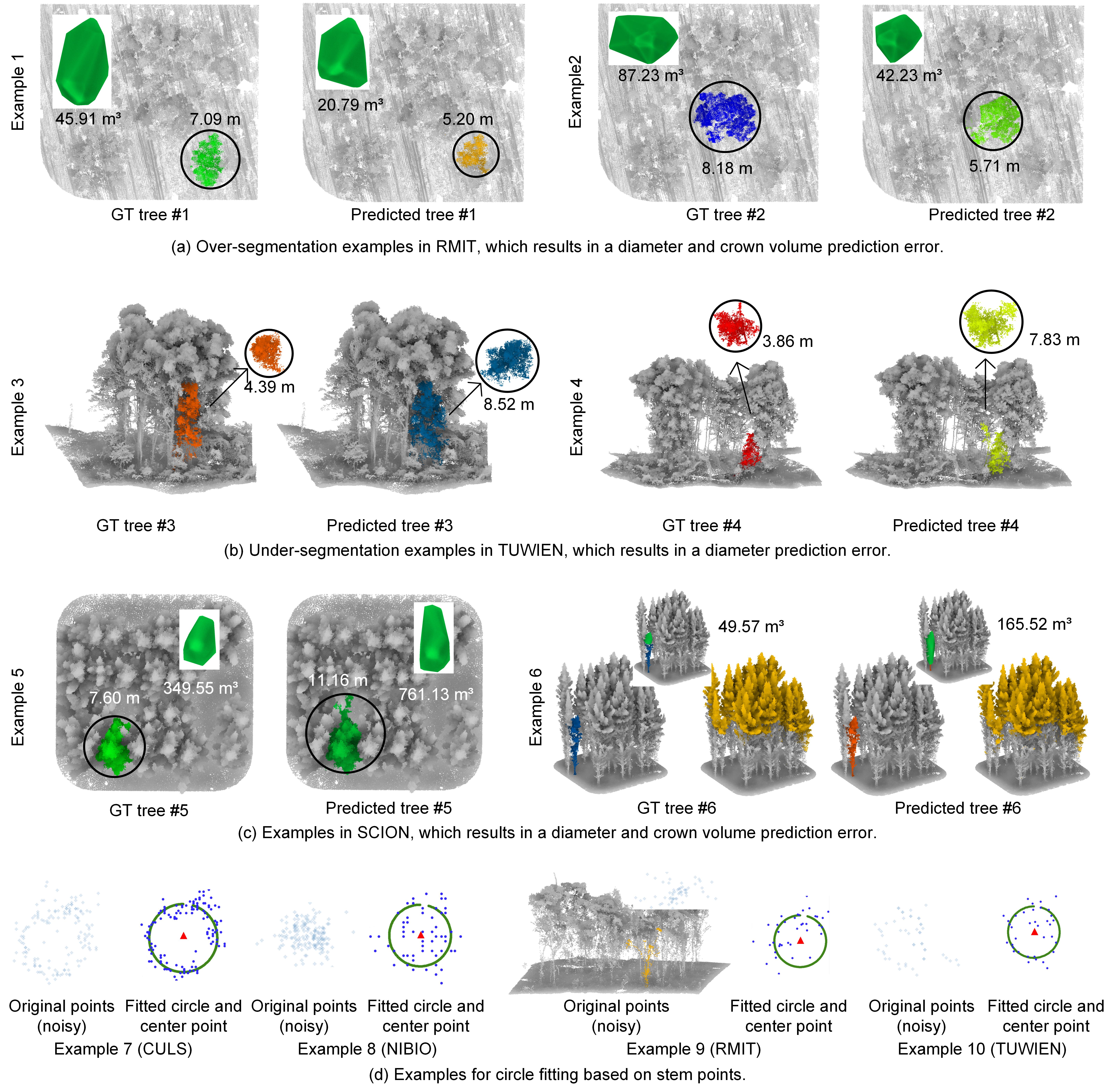}
\caption{Visual comparison of estimated and ground truth tree attributes.}
\label{fig:error_analysis_attributes}
\end{figure}

For the prediction of tree crown attributes, i.e., crown diameter, crown volume, and crown volume of live branches, both accurate individual tree segmentation and individual tree component classification are required. For crown diameter prediction, RMIT and TUWIEN have relatively worse predictions 0.26 RMSE\% for RMIT and 0.20 for TUWIEN, respectively. For example, two large, umbrella-shaped tree crowns in RMIT have been over-segmented, leading to incorrect crown dimaeter estimates, see Figure~\ref{fig:error_analysis_attributes}(a). Two examples from TUWIEN can be seen in Figure~\ref{fig:error_analysis_attributes}(b), where under-segmentation has\replaced[id=R1]{ led}{ lead} to large differences between estimated values and the reference. As the tree crowns in TUWIEN are large, inaccurately segmenting them causes significant mis-estimation of the crown diameter. Crown volumes (both over all branches and over only live branches) estimates are relatively poor for the RMIT and SCION regions. As shown in Figure~\ref{fig:error_analysis_attributes} (examples 1 and 2), the estimates are again too low due to over-segmentation. For SCION (example 5) the predicted value is too large due to under-segmentation. In addition, tree component segmentation has a significant influence on the results: although the individual tree segmentation is good, some dead branches are mis-classified as live branches, leading to the final crown volume of live branches being greatly over-estimated. This exposes a conceptual issue of this common variable: the crown volume of a tree has a considerable definition uncertainty, and a minor change in what is regarded as the crown can lead to large changes of the volume.

DBH and tree location are generally less accurate, which can be attributed to the sparsity and noise of the point clouds around the breast height that, in turn limit the accuracy of circle fitting. This is somewhat expected given the airborne recording geometry of the FOR-Instance data. Figure~\ref{fig:error_analysis_attributes}(d) shows some examples from each region. It can be seen that except for the CULS region, where the circle shape is relatively clear, the sparsity of the data makes accurate measurements difficult. Sometimes there may be no stem points at all, making it impossible to perform fitting. In those cases the average $x$, $y$ coordinates of all the points on a tree are used as its location. The sparsity of stem points highlights a limitation of ALS data. A solution could be to adopt a different method of estimating DBH, for instance via airborne allometric models~\citep{Jucker2017Allometric}. 

\begin{table}[]
\centering
\caption{Results for individual trees features and stand-wise attributes, for each plot in the test set.}
\label{tab:attributesPerPlot}
\resizebox{\textwidth}{!}{%
\begin{tabular}{lcrrrrccccrrrrr}
\hline
\multirow{3}{*}{\begin{tabular}[c]{@{}c@{}}Plot ID\\ (Region)\end{tabular}} &
  \multirow{3}{*}{\begin{tabular}[c]{@{}c@{}}Stand density\\ (trees/ha)\end{tabular}} &
  \multicolumn{2}{c}{DTM} &
  \multicolumn{2}{c}{Tree height} &
  \multicolumn{2}{c}{DBH RMSE} &
  \multicolumn{2}{c}{Crown diameter} &
  \multicolumn{2}{c}{\begin{tabular}[c]{@{}c@{}}Crown volume RMSE ($m^3$)\\ (RMSE\%)\end{tabular}} &
  \multicolumn{3}{c}{\begin{tabular}[c]{@{}c@{}}Location RMSE\\ (cm)\end{tabular}} \\ \cmidrule(l){3-4} \cmidrule(l){5-6}\cmidrule(l){7-8}\cmidrule(l){9-10}\cmidrule(l){11-12}\cmidrule(l){13-15}
\multicolumn{1}{c}{} &
   &
  \begin{tabular}[c]{@{}c@{}}Cov\\ (\%)\end{tabular} &
  \multicolumn{1}{c}{\begin{tabular}[c]{@{}c@{}}RMSE\\ (cm)\end{tabular}} &
  \begin{tabular}[c]{@{}c@{}}RMSE\\ (m)\end{tabular} &
  \multicolumn{1}{c}{RMSE\%} &
  \begin{tabular}[c]{@{}c@{}}With GT\\ (cm)\end{tabular} &
  \begin{tabular}[c]{@{}c@{}}With field data\\ (cm)\end{tabular} &
  \begin{tabular}[c]{@{}c@{}}RMSE\\ (m)\end{tabular} &
  \multicolumn{1}{c}{RMSE\%} &
  \multicolumn{1}{c}{Live branches} &
  All branches &
  x &
  \multicolumn{1}{c}{y} &
  \multicolumn{1}{c}{mean} \\ \hline
1 (CULS)  & 2079 & 100.0   & 1  & 0.5  & 0.02 & 0  & 5  & 0.1 & 0.01 & 13.1 (0.08)  & 14.3 (0.07)  & 0   & 0   & 0   \\
2 (NIBIO)  & 4771 & 98.8 & 7  & 0.6  & 0.02 & 6  & 7  & 0.5 & 0.10 & 45.8 (0.32)  & 47.6 (0.23)  & 29  & 11  & 20  \\
3 (NIBIO) & 4572 & 99.2 & 12 & 0.3  & 0.01 & 4  & 5  & 0.7 & 0.14 & 30.6 (0.34)  & 30.9 (0.19)  & 3   & 2   & 2   \\
4 (NIBIO) & 4612 & 99.6 & 12 & 0.8  & 0.03 & 7  & 9  & 0.4 & 0.08 & 25.6 (0.33)  & 16.5 (0.13)  & 8   & 31  & 19  \\
5 (NIBIO) & 3075 & 99.1 & 13 & 0.1  & 0.01 & 4  & 5  & 0.6 & 0.10 & 39.3 (0.26)  & 24.7 (0.11)  & 37  & 45  & 41  \\
6 (NIBIO) & 4015 & 99.0 & 15 & 0.9 & 0.03 & 4  & 5  & 0.6 & 0.15 & 27.0 (0.28)  & 32.2 (0.20)  & 3   & 2   & 3   \\
7 (NIBIO) & 3012 & 99.4 & 10  & 0.4  & 0.01 & 13 & 10  & 0.6 & 0.10 & 55.4 (0.33)  & 47.1 (0.22)  & 68  & 38  & 53  \\
8 (RMIT) & 7283 & 100.0   & 3  & 0.5  & 0.06 & 16 & 19 & 0.7 & 0.26 & 9.2 (0.96)   & 10.0 (0.94)   & 36 & 39 & 37 \\
9 (SCION) & 2758 & 99.9 & 3  & 0.1  & 0.00 & 6  & 12 & 1.0 & 0.14 & 81.4 (0.35)  & 115.1 (0.37) & 7   & 6   & 6   \\
10 (SCION) & 2273 & 100.0   & 4  & 0.0  & 0.00 & 14 & 11 & 0.9 & 0.11 & 114.9 (0.38) & 242.8 (0.68) & 151 & 58  & 104 \\
11 (TUWIEN) & 3419 & 100.0 & 26 & 0.4  & 0.02 & 27 & 26 & 1.4 & 0.20 & 137.8 (0.60) & 142.2 (0.57) & 117 & 55  & 86  \\ \hline
\end{tabular}%
}
\end{table}

\section{Discussion}
\label{Sec:Discussion}

\subsection{Baseline comparisons}
\label{Sec:EffectivenessofSegmentationComparedWithOtherMethods}

We have compared our proposed pipeline to three alternatives. Our first baseline\replaced[id=R1]{ is an example of current standard}{ is the best} practice, available in popular cloud processing software: Treeiso~\citep{Xi2022Dec2Graph} is a recently published, unsupervised algorithm to segment individual trees, which is available as a plug-in for the CloudCompare software (\href{https://www.cloudcompare.org}{https://www.cloudcompare.org}, last accessed 09/2023).
Treeiso requires point clouds consisting exclusively of tree points. We therefore feed it only the points assigned to  trees by our semantic segmentation module, under the TreeMix setting. The user parameters of Treeiso must be tuned for optimal performance, but are quite user-friendly and understandable also for non-experts in forestry. 

The method consists of three steps. For the first step, an initial segmentation with a 3D cut-pursuit algorithm, we found the default values to be effective. In the second and third steps, there are three parameters that influence the segmentation results, as shown in Table~\ref{tab:bestParasTreeiso}: the number $K$ of nearest neighbors to search in step 2, a 2D cut-pursuit algorithm; the strength $\lambda$ of the regularization in that step; and the height/length ratio $Th_h$ for step 3, a global refinement. Larger values of $K$ and $\lambda$, and smaller values of $Th_h$, result in larger segments. I.e., $K$ and/or $\lambda$ must be decreased or $Th_h$ must be increased in case of under-segmentation, and vice versa if over-segmentation occurs.

Table~\ref{tab:comparewithTreeiso} shows the best individual tree segmentation accuracy that we could achieve with careful tuning. We note that, due to the interactive procedure, it is impossible to exhaust all parameter combinations. We started from the default values and iteratively adjusted them ($K$ and $\lambda$ in steps of 5, $Th_h$ in steps of 0.05) to obtain the visually best output. In this way we managed to gain almost 8\,pp in F-score compared to the default parameters, see Table~\ref{tab:comparewithTreeiso}. Still a significant gap ($\approx$17\,pp) remains compared to the performance of our deep learning method.

Figure~\ref{fig:semanticandinstanceseg} shows a qualitative comparison between Treeiso and our segmentation method. In the CULS region, both algorithms achieve excellent individual tree segmentation results. In all other regions we find that the deep learning-based methods has advantages over Treeiso in certain cases (marked by white ellipses). For plots where trees vary greatly in size, such as RMIT and TUWIEN, we could not find a way to tune the parameters of Treeiso such that it correctly segments trees of different sizes within the plot. It either over-segments large trees or merges small trees, whereas the learned segmentation model handles these situations better. In addition, for coniferous trees with jagged shapes such as those in the NIBIO and SCION examples, parameter tuning is also difficult to find suitable parameters for handling tightly spaced trees. Especially protruding branches are over-segmented, and none of the parameter setting we tried could rectify this.

\begin{table}[]
\centering
\caption{\added[id=R1]{Quantitative comparison to individual tree segmentation with Treeiso. All values are percentages (\%). ``default'' are results obtained with the recommended default parameters of Treeiso, while ``best'' are those obtained with parameters that we individually tuned for each plot. Refer to Table~\ref{tab:bestParasTreeiso} for those optimal values.}}
\label{tab:comparewithTreeiso}
\small 
\setlength{\tabcolsep}{3pt} 
\renewcommand{\arraystretch}{0.8} 
\begin{tabular}{@{}lccccc@{}}
\toprule
Method                       & Completeness & Omission error & Commission error & F-score & Cov\\ \midrule
Treeiso (default) & 64.1        & 35.9          & 43.1           & 60.3   &    67.8\\
Treeiso (best)    & 70.0        & 30.0          & 33.5           & 68.2   &   71.7\\
Ours                         & \textbf{82.0}        & \textbf{18.0}          & \textbf{11.7}           & \textbf{85.1}   &  \textbf{78.1}\\ \bottomrule
\end{tabular}%
\end{table}

\begin{table}[]
\centering
\caption{Default and best Treeiso parameter combinations for each plot in FOR-Instance.}
\label{tab:bestParasTreeiso}
\resizebox{\textwidth}{!}{%
\begin{tabular}{@{}lcccccccccccc@{}}
\toprule
Plot ID & 1 & 2 & 3 & 4 & 5 & 6 & 7 & 8 & 9 & 10 & 11 & Default \\ \midrule
$K$ (Nearest neiborhood to search in step 2: 2D cut-pursuit algorithm) & 50  & 20  & 20  & 10  & 15  & 20  & 15  & 20  & 20  & 20   & 20  & 20  \\
$\lambda$ (Regularization strength in step 2)           & 20  & 10  & 25  & 20  & 10  & 10  & 15  & 10  & 20  & 20   & 40  & 20  \\
$Th_h$ (Relative height length ratio in step 3: global refinement)    & 0.5 & 0.3 & 0.5 & 0.5 & 0.3 & 0.5 & 0.3 & 0.7 & 0.6 & 0.05 & 0.5 & 0.5 \\ \bottomrule
\end{tabular}%
}
\end{table}

Another comparison was made against a different deep learning-based method~\citep{Straker2023Instance}. That method involves constructing a CHM from the point cloud and converting it to pseudo-colour images\added[id=R1]{ of 640 $\times$ 640 pixels}, which are then fed into a 2D object detection network, YOLOv5~\citep{Jocher2022yolov5}.\added[id=R1]{ They used the published weights of the YOLOv5l-seg model and chose the same hyperparameters as described by~\cite{Jocher2022yolov5}, except for changing the initial learning rate to 0.001.} The quantitative comparison is given in Table~\ref{tab:compareYOLOv5}. Our segmentation-based approach achieved higher F-scores in all forest regions except for CULS. Consistent with the findings reported by ~\citet{Straker2023Instance}, both methods reach higher quality in coniferous forests, such as CULS, SCION, and NIBIO; whereas the quality somewhat deteriorates in forests with strongly varying tree heights and crown sizes, as in the RMIT and TUWIEN. Still, also in those challenging regions our segmentation method had an edge. On the particularly challenging TUWIEN plot that more than doubles the F-score from 30\% to 69.4\%.

Finally we also compare our individual tree segmentation results\deleted[id=R1]{ with} to those of Point2Tree~\citep{Wielgosz2023Jul27P2T}, another graph-based\added[id=R1]{ method} that was originally developed for terrestrial mobile laser scanning (MLS) point clouds. Point2Tree consists of two stages. First it employs a PointNet++ model to obtain semantic labels for the points. The semantic segmentation serves as a basis for an unsupervised, graph-based algorithm. That instance segmentation algorithm sequentially identifies individual trees in a graph: the segmented point cloud is divided into small segments that form the nodes of the graph, constructed based on trunk information. For optimal results, various hyper-parameters must be selected individually for each forest type, including slice size, minimum number of stem points, and stem height.

Point2Tree is particularly sensitive to low point density near the ground. This sensitivity arises because, having been developed for TLS points, graph construction starts from the bottom, and the PointNet++ has been trained on MLS data. If the point cloud is too sparse there, the method tends to break, which makes it less suitable for our ALS data. For instances with adequate density at lower height, Point2Tree is effective. When applied to FOR-Instance, the method was only successful for 3 of the test plots, shown in Table~\ref{tab:compareP2T}\added[id=R1]{; showing that our method, adapted to the characteristics of ALS data, indeed offers a significant advantage when it comes to handling plots with varying point densities. For completeness} we note that the failure cases could possibly be alleviated by re-training Point2Tree for ALS data and then adapting the graph -based part to the new data characteristics.

\begin{table}[]
\centering
\caption{Quantitative comparison to CHM-based YOLOv5~\citep{Straker2023Instance} on the individual tree segmentation task. All values are percentages ($\%$).}
\label{tab:compareYOLOv5}
\small 
\setlength{\tabcolsep}{3pt} 
\renewcommand{\arraystretch}{0.8} 
\begin{tabular}{@{}llrrrr@{}}
\toprule
Region                  & Method & Completeness & Omission error & Commission error & F-score        \\ \midrule
\multirow{2}{*}{CULS}   & YOLOv5 & 100\textcolor{white}{.0}          & 0\textcolor{white}{.0}              & 0\textcolor{white}{.0}                 & \textbf{100}\textcolor{white}{.0}   \\ 
                        & Ours   & 100.0        & 0.0            & 13.0             & 93.0          \\ \midrule
\multirow{2}{*}{NIBIO}  & YOLOv5 & 72\textcolor{white}{.0}            & 28\textcolor{white}{.0}              & 13\textcolor{white}{.0}                & 79\textbf{\textcolor{white}{.0}}              \\
                        & Ours   & 88.4        & 11.6          & 3.6             & \textbf{92.4} \\ \midrule
\multirow{2}{*}{RMIT}   & YOLOv5 & 62\textcolor{white}{.0}            & 38\textcolor{white}{.0}              & 30\textcolor{white}{.0}                & 65\textbf{\textcolor{white}{.0}}              \\
                        & Ours   & 64.1         & 35.9           & 24.1             & \textbf{69.5} \\ \midrule
\multirow{2}{*}{SCION}  & YOLOv5 & 91\textcolor{white}{.0}            & 9\textcolor{white}{.0}               & 9\textcolor{white}{.0}                 & 91\textbf{\textcolor{white}{.0}}              \\
                        & Ours   & 87.7         & 12.3           & 4.0              & \textbf{91.5} \\ \midrule
\multirow{2}{*}{TUWIEN} & YOLOv5 & 23\textcolor{white}{.0}            & 77\textcolor{white}{.0}              & 59\textcolor{white}{.0}                & 30\textbf{\textcolor{white}{.0}}              \\
                        & Ours   & 71.4         & 28.6           & 33.4             & \textbf{69.4} \\ \bottomrule
\end{tabular}%
\end{table}

\begin{table}[]
\centering
\caption{Quantitative comparison to Point2Tree~\citep{Wielgosz2023Jul27P2T} on the individual tree segmentation task. Point2Tree was only successful on three of the test plots. All values are percentages ($\%$).}
\label{tab:compareP2T}
\small 
\setlength{\tabcolsep}{3pt} 
\renewcommand{\arraystretch}{0.8} 
\begin{tabular}{@{}llcccc@{}}
\toprule
\begin{tabular}[c]{@{}l@{}}Plot ID\\ (Region)\end{tabular} & Method & Completeness & Omission error & Commission error & F-score \\ \midrule
\multirow{2}{*}{CULS}      & Point2Tree  & 93.4   & 6.6   & 48.8    & 61.5      \\
                           & Ours   & 100.0       & 0.0   & 13.0    & \textbf{93.0} \\ \midrule
\multirow{2}{*}{NIBIO plot 3} & Point2Tree  & 76.0 & 24.0 & 68.5 & 40.0          \\
                           & Ours & 96.7 & 3.3  & 0.0     & \textbf{98.3} \\ \midrule
\multirow{2}{*}{NIBIO plot 4} & Point2Tree  & 80.5 & 19.5 & 64.9 & 47.2          \\
                           & Ours & 96.3 & 3.7  & 0.0     & \textbf{98.1} \\ \bottomrule
\end{tabular}%
\end{table}

\subsection{Dominant vs.\ understory trees}
\label{Sec:ConsideringDominantAndSmallerTrees}

\begin{figure}
\centering
\includegraphics[width=\textwidth,keepaspectratio]{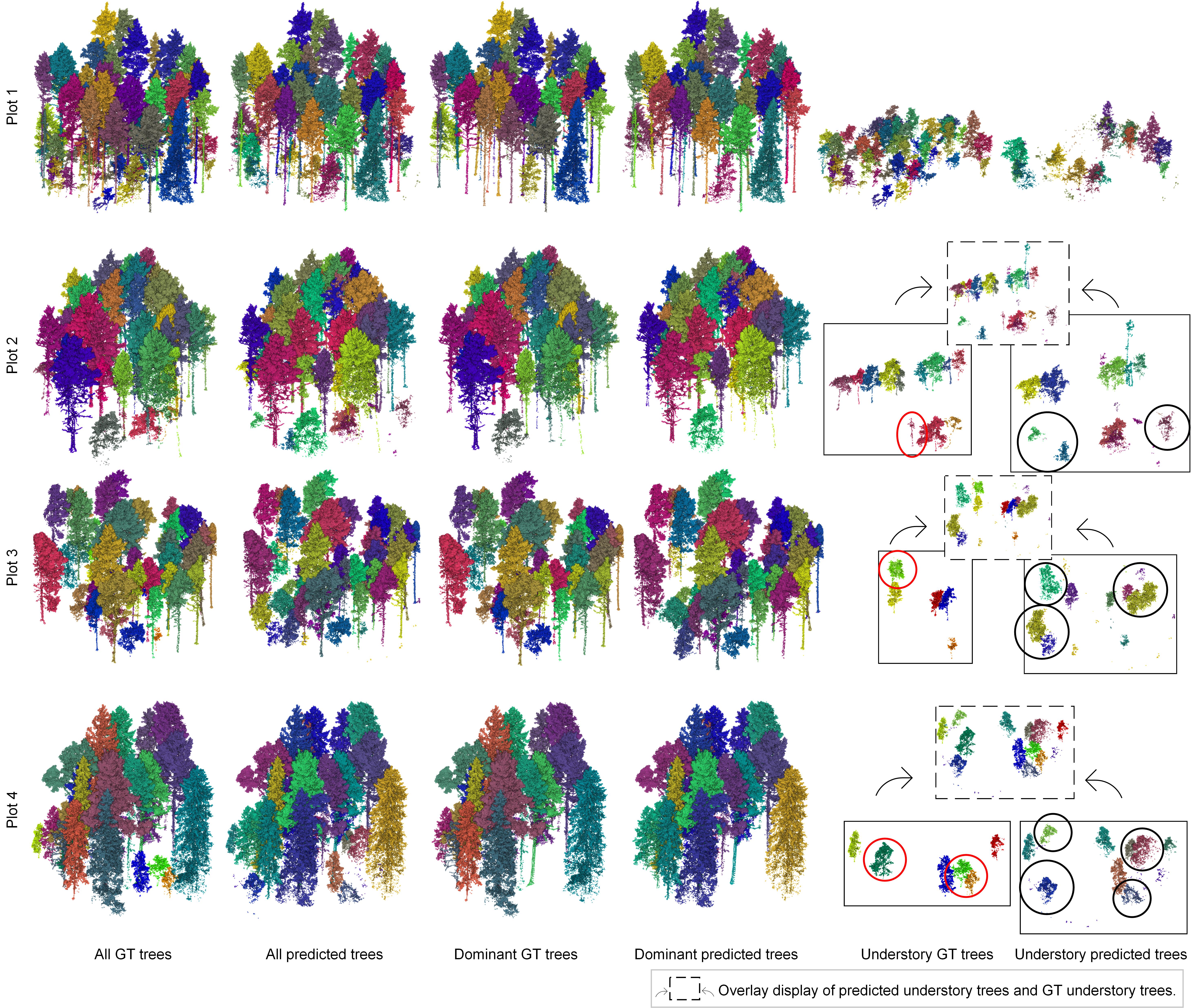}
\caption{Qualitative examples of individual tree segmentation in the NIBIO2 region. Dominant trees are estimated rather well, whereas the distinction between understory trees and low vegetation is somewhat ambiguous.}
\label{fig:dominant_trees}
\end{figure}

Besides the 11 plots in the test set, the FOR-Instance data includes another 15 plots also collected in Norway, which we refer to as NIBIO2. The data from NIBIO2 have not been included in the test set for two reasons. First, NIBIO2 does not have reference measurements of DBH. Second, in this area the reference data exhibits evident confusion between the low vegetation class and understory trees: many small trees are not annotated in the reference, which would introduce a bias in the quantitative results. 
We therefore separately evaluate on NIBIO2, using 1/3 of the maximum tree height as the threshold to separate dominant from understory trees, as  described in Section~\ref{Sec:MetricsForConsideringSmallerTrees}. Table~\ref{tab:dominantTrees} and Figure~\ref{fig:dominant_trees} present the corresponding results for NIBIO2.

As shown in Table~\ref{tab:dominantTrees}, the performance metric are much lower across all trees than when measured only across the dominant trees (72.8$\%$ vs.\ 82.9$\%$ in F-score). When measured only across understory trees, the F-score drops to 39.3$\%$, showing their substantial impact on the aggregate metrics. As can be seen in  Figure~\ref{fig:dominant_trees}, the gap results from trees that are (correctly) identified by the network, but not annotated in the reference, resulting in commission errors that are, in fact, due to limitations of the reference data. In the last two columns of Figure~\ref{fig:dominant_trees}, understory trees that are present in the reference but missed by the segmentation network are circled in red, whereas those found by the network but missing in the reference are circled in black. It is clear that the latter are plausible tree detections that should not be penalised.\added[id=R1]{ This discrepancy highlights the need for more complete ground truth annotations, especially for understory trees, to properly assess tree segmentation.

Our method treats dominant and understory trees in a unified manner, in principle ensuring equal effectiveness for both. Still, understory trees may exhibit lower point density and suffer from missing data caused by severe occlusions, due to the nature of airborne sensor data acquisition. Consequently, their segmentation is usually less accurate  compared to canopy-level trees~\citep{Dong2020Multilayered,Jarron2020Detection,Wang2023Automatic}. We are unable to quantify the difference in segmentation performance between dominant and understory trees for our pipeline, because of the lack of precise ground truth annotations in the dataset. This uncertainty again emphasizes the need for more complete and accurate ground truth. Reference datasets with carefully annotated understory trees will be a indispensable to overcome the current limitations and unlock the full potential of segmentation algorithms for forestry.}

\begin{table}[h]
\centering
\caption{Quantitative results of individual tree segmentation in the NIBIO2 region. One can clearly see the effect of the understory, and associated labeling ambiguities, on the performance metrics.}
\label{tab:dominantTrees}
\resizebox{\textwidth}{!}{%
\begin{tabular}{@{}lcccccc@{}}
\toprule
Groups &
  \begin{tabular}[c]{@{}c@{}}reference\\ trees\end{tabular} &
  \begin{tabular}[c]{@{}c@{}}detected\\ trees\end{tabular} &
  \begin{tabular}[c]{@{}c@{}}correctly detected trees\\ (Completeness)\end{tabular} &
  \begin{tabular}[c]{@{}c@{}}omitted trees\\ (Omission \replaced[id=R1]{error}{ rate})\end{tabular} &
  \begin{tabular}[c]{@{}c@{}}wrong detections\\ (Commission \replaced[id=R1]{error}{ rate})\end{tabular} &
  F-score \\ \midrule
Dominant &
  571 &
  490 &
  440 (77.1\%) &
  131 (22.9\%) &
  50 (10.2\%) &
  82.9\% \\
Understory trees&
  266 &
  157 &
  83 (31.2\%) &
  183 (68.8\%) &
  74 (47.1\%) &
  39.3\% \\
All trees&
  837 &
  647 &
  540 (64.5\%) &
  297 (35.5\%) &
  107 (16.5\%) &
  72.8\% \\ \bottomrule
\end{tabular}%
}
\end{table}

\subsection{Implications and limitations}
\label{Sec:ImplicationAndLimitation}

The pipeline proposed in this paper sets a new state of the art for automated individual tree segmentation and semantic segmentation in ALS point clouds (at least for the forest characteristics covered by FOR-Instance\added[id=R1]{, including boreal coniferous forest, temperate coniferous forest, temperate deciduous forest and sclerophyll forest}).

\added[id=R1]{ Our method has also demonstrated potential for detecting understory trees. While contributing minimally to biomass, they play a crucial role for the ecosystem by ensuring canopy succession, stand development, soil erosion protection, and habitat for wildlife~\citep{Hamraz2017Forest}. The detection of understory trees has recently received more attention within the research community~\citep{Hamraz2017Forest,Dong2020Multilayered,Jarron2020Detection,Wang2023Automatic,Penner2023Automated}. Current methods often follow a stratified approach that iteratively removes higher canopy layers, then detects lower ones~\citep{Dong2020Multilayered,Jarron2020Detection,Wang2023Automatic}. In contrast, our method directly processes 3D point clouds, without the need to derive a 2.5D height model and to update it for each layer. We argue that directly analyzing the 3D forest structure is a more natural approach to effectively and efficiently detect understory trees. Moreover, advances in LiDAR technology (e.g., single-photon LiDAR) can be expected to further increase point cloud density and to promote the adoption of 3D methods.}

Based on the segmentation, it also delivers accurate estimates of key variables like tree height, crown width, and crown volume, as well as stand attributes like stem density and the terrain of the forest floor. These outputs provide a strong basis for a tree-level forest inventory, and consequently for fine-scale forest management. Complete per-tree information opens up the possibility to monitor tree growth dynamics over extended areas. It may also contribute to more targeted timber production, safeguarding of young trees, and maintenance of biodiversity within forest ecosystems. Moreover, the segmentation results may enhance the automatic estimation of ecosystem service indicators as well as aggregate timber volume, carbon stock and sequestration, in support of ecological management and protection.

\begin{figure}
\centering
\includegraphics[width=\textwidth,keepaspectratio]{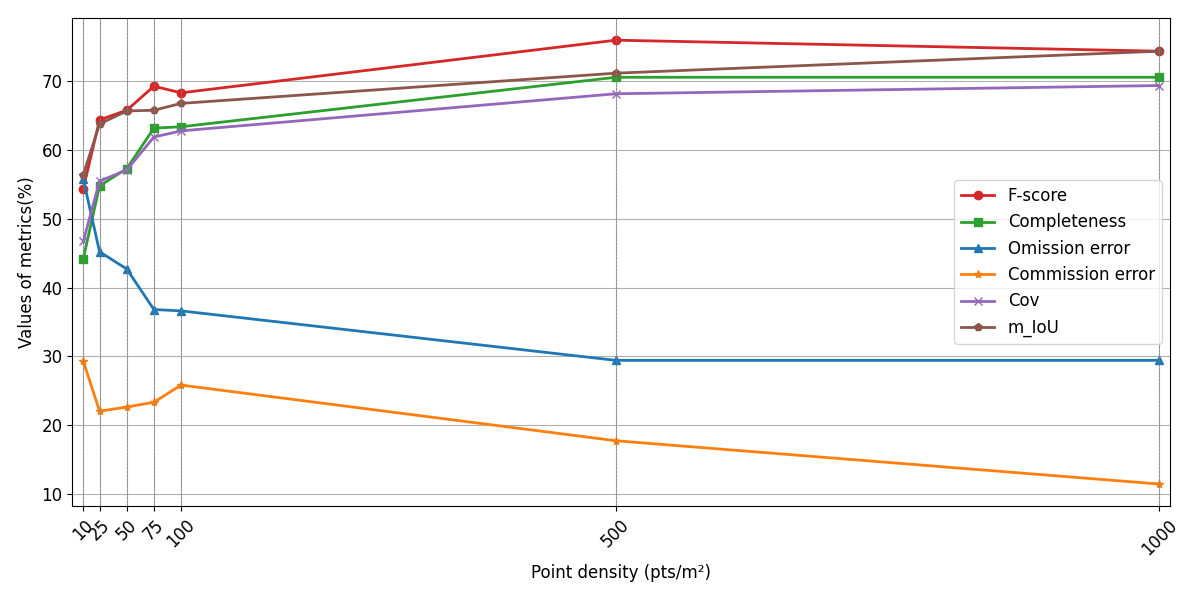}
\caption{\added[id=R1]{Variations in segmentation metrics across point densities.}}
\label{fig:density_experiments}
\end{figure}

Of course, our proof-of-concept also has limitations.\added[id=R1]{ In particular, it tends to reach lower segmentation accuracy for more complex forest structure. Among the forest types contained in the FOR-Instance dataset, the CULS region consists solely of even-aged trees of a single tree species (Scots pine). The NIBIO area also features mature forest, dominated mainly by Picea abies. Although the SCION region has high tree density ($\approx$930 trees per hectare), it is a productive plantation forest also dominated by a single species (Pinus radiata). The segmentation accuracy in these three regions is high, which can be attributed to their relatively uniform species composition, with even-aged and mature trees creating a regular, single-layered forest structure. Despite the overlapping and intertwined crowns of the closely spaced, radiata pines with many branches, the segmentation accuracy in the SCION area remains high.

On the contrary, the TUWIEN and RMIT regions are structurally more complex. TUWIEN features complicated vegetation structures (woody debris, lying and standing deadwood, and dense understory), a large diversity of tree species, and considerable variations in canopy layering and tree size (DBH up to 1.3\,m). RMIT encompasses trees with a wide range of ages and heights (from 5\,m to 17\,m), with an understory consisting of low shrubs ($\leq$2\,m in height) and native grasses. For these two structurally more intricate forests our method exhibits markedly lower accuracy (Figure~\ref{tab:IndividualTreeSegBest}). It is worth noting again that TUWIEN and RMIT both only have a single training plot in the current training data. It is thus not clear to what extent the lower performance stems from the higher complexity and not simply from insufficient training data. It will be interesting to see how 3D deep learning copes with more varied forests once sufficiently large training sets become available.

Another factor is point cloud density. We have conducted a simple experiment to study the impact of point density on the segmentation accuracy of ForAINet: The original FOR-Instance data is sub-sampled to 7 different point densities (i.e., 10, 25, 50, 75, 100, 500, and 1000\,pts/m\textsuperscript{2}). That range of densities encompasses the typical densities of our ALS-HD (500-1000\,pts/m\textsuperscript{2}), through conventional, airborne high-density ALS (up to 100\,pts/m\textsuperscript{2}), to traditional ALS (approximately 10\,pts/m\textsuperscript{2}). We then run our best configuration (with TreeMix augmentation, see Section~\ref{Sec:3DPointCloudSegmentation}) on the sub-sampled datasets. The results indicate that the performance decreases with point density (Figure~\ref{fig:density_experiments}). The drop in instance segmentation F-score remains within 5 percentage points for  densities $\geq$75\,pts/m\textsuperscript{2}. Below that value the omission error markedly increases. The mIoU for semantic segmentation follows a similar pattern. In conclusion, our 3D learning method is challenged by point densities below 100\,pts/m\textsuperscript{2}, as obtained for instance from traditional airborne ALS.}

\replaced[id=R1]{Another}{ One} important point is that \replaced[id=R1]{our method}{ it} builds on fully supervised machine learning. Its performance is strongly heavily by the data seen during training, which means that it cannot necessarily be deployed out-of-the-box to unfamiliar forest types\added[id=R1]{ (e.g., tropical environments with high overlap between trees)} or sensor characteristics. Empirically, even moderate changes of the data characteristics can greatly affect neural network estimates. To sidestep extensive data collection efforts it may be useful to explore transfer learning and domain adaptation strategies~\citep{Triess2021Survey}. To that end, further empirical studies are needed, which in turn call for even larger and more diverse reference datasets.

Our current experiments are restricted to ALS-HD data. While that acquisition setup proved to be an excellent choice not only for tree height, but also for and crown-related attributes, retrievals of DBH and tree location were less accurate. The circle fitting used for those tasks is compromised by the lower point density of ALS on lower parts of the stem. We see two potential solutions: (i) The use of multi-source LiDAR, by complementing ALS with terrestrial (possibly mobile) scans. we are confident that this could largely resolve the problem, but it would come at the cost of a larger scanning effort, and poses the additional technical challenges to co-register terrestrial and airborne data to centimeter level. (ii) A practical perhaps more attractive solution could be to avoid explicit, geometric estimation of stem parameters, and instead retrieve DBH and tree location directly from the ALS data. This could potentially be done either by augmenting the segmentation network with an appropriate regression branch, or by deriving dedicated allometric relations.  

\section{Conclusion}
\label{Sec:Conclusions}

We have assembled a point cloud processing pipeline aimed at extracting a complete inventory of per-tree attributes from ALS-HD data. The pipeline starts by extracting semantic class labels and individual trees \added[id=R1]{with ForAINet}, which then facilitate the automatic estimation of tree- and plot-level attributes. In experiments on the FOR-Instance dataset the proposed method achieved excellent segmentation performance, with 85.1$\%$ F-score for individual tree segmentation and a 73.5$\%$ mIoU for 5-class semantic segmentation. The good segmentation quality translates to good predictive skill for several important biophysical properties, including tree height, crown width, crown volume. Retrievals of DBH and tree location were less accurate, because the scanning geometry of ALS results in an unfavourable point density on the lower stems.\added[id=R1]{ Overall, the proposed method shows good performance across a variety of forest types, including the detection of many understory trees. Segmentation quality does, however, noticeably deteriorate if a forest exhibits complex structure, and also if the point density falls below $\approx$100\,pts/m$^2$.} While further research is needed to address these remaining issues, we believe that fully automated tree-level forest inventories based on remotely sensed data are within reach.


\small
\bibliographystyle{elsarticle-harv}
\bibliography{rsebib-forestryinventory}

\clearpage
\appendix
\section{List of abbreviations}

\begin{table}[!h]
\centering
\caption{}
\label{tab:abbreviations}
\small  
\begin{tabular}{@{}ll@{}}
\toprule
Abbreviations and symbols & Meaning \\ \midrule
ABA   & area-based approach \\
ALS   & airborne laser scanning \\
ALS-HD  & very high density ALS point clouds\\
CHM   & canopy height model \\
CWD   & coarse woody debris \\
DBH   & diameter at breast height \\
DEM   & digital elevation model \\
DLS   & drone laser scanning \\
DSM   & digital surface model \\
DTM   & digital terrain model \\
FCN   & fully convolutional network\\
GT   & ground truth \\
GUI   & graphical user interface \\
CV   & computer vision \\
IoU   & intersection over union \\
ITC   & individual tree crown \\
ITD   & individual tree detection \\
LiDAR & Light Detection and Ranging \\
MLP   & multi-layer perception \\
MLS   & mobile laser scanning \\
NMS   & non-maximum suppression \\
RMSE  & root mean square error\\
TLS   & terrestrial laser scanning \\
UAV   & unmanned aerial vehicle \\
ULS   & unmanned laser scanning \\\bottomrule
\end{tabular}
\end{table}

\end{document}